\documentclass[twocolumn]{IEEEtranTCOM}
\usepackage[dvips]{graphicx}
\graphicspath{{../eps/}{../pdf/}{../jpeg/}}
\DeclareGraphicsExtensions{.eps,.pdf,.jpeg,.png}
\usepackage[cmex10]{amsmath}
\usepackage{xcolor,amsthm, amsfonts, mdwmath, mdwtab, eqparbox, psfrag, setspace}
\usepackage[tight,footnotesize,sf,SF]{subfigure} 
\usepackage{cite,multicol, blindtext,overpic,url,color,tikz,lipsum}
\usepackage{bbm}
\definecolor{dg}{rgb}{0.1,0.55,0.15}

\usepackage{algpseudocode,algorithm}
\newtheorem{proposition}{Proposition}
\newtheorem{lemma}{Lemma}

\usepackage[pass]{geometry}

\newtheorem{remark}{Remark}
\newtheorem{corollary}{Corollary}
\usepackage{cancel,fancyhdr,acronym}
\usepackage[T1]{fontenc}
\usepackage[normalem]{ulem}
\usepackage{soul}

\newcommand\bout{\bgroup\markoverwith{\textcolor{blue}{\rule[.5ex]{2pt}{1.0pt}}}\ULon}
\definecolor{dg}{rgb}{0.48, 0.25, 0.0}

\usepackage{subfigure}
\usepackage{authblk}
\definecolor{seagreen}{rgb}{0.18, 0.55, 0.34}

\begin{document}
\title{Learning-based decentralized offloading decision making in an adversarial environment}	
\author{
	\IEEEauthorblockN{Byungjin Cho and Yu Xiao} 
%
	\vspace*{-.5cm}	
%
%
%
	\thanks{This work has received funding from the European Union's Horizon 2020 research and innovation programme under grant agreement No. 825496 and No. 815191, and Academy of Finland under grant number 317432 and 318937.
	
	B. Cho and Y. Xiao are with the Department of Communications and Networking, Aalto University, 00076 Espoo, Finland (e-mail: byungjin.cho@aalto.ﬁ; yu.xiao@aalto.ﬁ).
	
}}

\maketitle
\begin{abstract}
	Vehicular fog computing (VFC) pushes the cloud computing capability to the distributed fog nodes at the edge of the Internet, enabling compute-intensive and latency-sensitive computing services for vehicles through task offloading. However, a heterogeneous mobility environment introduces uncertainties in terms of resource supply and demand, which are inevitable bottlenecks for the optimal offloading decision. Also, these uncertainties bring extra challenges to task offloading under the oblivious adversary attack and data privacy risks.  
	{In this article, we develop a new adversarial online learning algorithm with bandit feedback based on the adversarial multi-armed bandit theory, to enable scalable and low-complexity offloading decision making. Specifically, we focus on optimizing fog node selection with the aim of minimizing the offloading service costs in terms of delay and energy. The key is to implicitly tune the exploration bonus in the selection process and the assessment rules of the designed algorithm,} taking into account volatile resource supply and demand. We theoretically prove that the input-size dependent selection rule allows to choose a suitable fog node without exploring the sub-optimal actions, and also an appropriate score patching rule allows to quickly adapt to evolving circumstances, which reduce variance and bias simultaneously, thereby achieving a better exploitation-exploration balance. Simulation results verify the effectiveness and robustness of the proposed algorithm. 
\end{abstract}

\section{{Introduction}}
 
{

Increasing demand for high-complexity but low-latency computation, triggered by emerging applications, e.g. autonomous driving, motivates the use of rising technologies, mobile edge/fog computing, that bring cloud-like computing services, closer to end-users \cite{Wang2017, Mao2017, Mouradian2018}. To boost up additional but {limited} edge computing resources, vehicular fog computing (VFC) \cite{Hou2016, Datta2017} has emerged as a new computing paradigm where moving fog nodes with surplus resources and good connectivity, named vehicular fog nodes (VFNs), are utilized as viable components that serve to execute computation tasks offloaded from service clients. As such, leveraging distributed fog nodes for task offloading could benefit from direct communication, e.g. 5G V2V, between a client and a VFN, i.e., reduced transmission delay, and similar trajectories when a client is traveling along with VFNs, i.e., relatively long contact duration and less handoffs, resulting in a substantial improvement in quality of experience, compared with using fixed infrastructure. VFNs are heterogeneous in terms of location, availability and reputation, and thus the computing service has diverse preferences towards them \cite{Chen2019}, i.e., one may prefer a vehicle with high processing capability and efficiency. One issue is how to make task offloading decisions, especially the fog node selection, considering distinct characteristics/preferences.
 
}

Computation task offloading decision algorithms have been investigated in \cite{Zheng2015, Gu2018} where a centralized coordinator schedules the computation offloading tasks. A decision-making problem has been formulated in \cite{Zheng2015} as a stochastic control process, e.g., semi-Markov, to minimize the offloading service cost in terms of delay and energy. The trade-off between the delay and energy cost is investigated in \cite{Gu2018} based on matching theory. However, such centralized decision-making might be challenging to run due to i) {signaling overhead burden} caused by gathering and processing a massive amount of information, e.g., requested tasks of service users, available resources of VFNs, and mobility of both, and ii) a {privacy concern} raised by exchanging such private information with a central controller.

Decentralized decision-making is considered as an alternative for the issues above. Each client can make a decision independently and perform task offloading in a distributed manner. A client may lack the state information of neighboring VFNs within its communication range, and thus it is unknown in prior that which VFN would provide the best performance, i.e., the lowest offloading cost. Exchanging the state between the client and the potential VFN, may be informative and helpful for making a decision appropriately. However, such decentralized decision-making is still challenging to conduct in a {mobile} environment where i) {frequent state updates} are needed to adapt to system fluctuation, and ii) such heavy signaling load could cause transmission failure and thus {outdated state}. One approach to deal with the issues above is, rather than obtaining the state information of VFNs from signaling messages, to enable a client to directly learn the state information of VFNs and to map the decision history to the current offloading decision.

Given the availability of a huge amount of data, historical data can be used to improve the quality of resource management policies, since they contain statistics of the environment which varies in a non-stationary and unknown manner, and learning from them can mitigate the uncertainty of future management tasks. Further capability to reinforce the current policies allows envisioning a learn-to-optimize framework where a decision is made in an environment to optimize a given notion of cumulative loss with the fewest possible assumptions. In a nutshell, the adaptive decision-making procedure becomes two-fold, i) exploration: learning as much as possible about different candidate actions that lead to good estimates of their loss, and ii) exploitation: optimizing the desired objective to select the optimal actions given the learned information.

One fundamental issue is {to balance the exploration and exploitation trade-off in the learning process}, i.e., making decisions with the aim of reducing uncertainty over states, or maximizing cumulative reward given its current estimates. Such an exploration versus exploitation dilemma can be formulated as a multi-armed bandit (MAB) problem where each neighboring VFN is treated as an independent arm, and its associated offloading service cost, e.g., latency and energy at edge node, is dominated by the computing capability. A task requester performs an online learning process, while running the offloading service and updating the optimal decision on the VFN selection. However, the variations in the requested workload (dynamic resource demand) and the candidate VFN set (dynamic resource supply) make it non-trivial for the task requester to learn the latency and energy consumption of the candidate VFNs especially in a rapidly changing or adversarial environment on which an attacker may give some stress.

In our task offloading problem, two dynamic factors are considered in terms of resource supply and demand: i) a time-varying volatile candidate fog node set which results from its inherent mobility \cite{Zhang2020}, and ii) a time-varying task size which results from different types of applications or even different parts of the same application workload \cite{Li2019}. Both factors above cause unnecessary but inevitable costs, i.e., sub-optimal actions are possibly explored thereby unbalancing exploration and exploitation activities. In particular, newly and re-appeared arms may fail to quickly adapt to the evolving circumstance. Also, a large-size task offloaded to a fog node with weak service capability may provoke worse performance. In the literature, such dynamic attributes have not been taken into account due to challenges associated with i) randomized selection rules and ii) unbiased estimation assessment rules in an adversarial environment, shed lighted in this work.

To the best of our knowledge, this is the first work aiming at bridging such dynamics to an adversarial domain, addressing the following contributions;
\begin{itemize}
	\item This work proposes a modified implicit exploration-based algorithm for adaptive learning-based task offloading (MIX-AALTO) which enables scalable and low-complexity decision making on VFN selection toward minimizing task offloading service cost in terms of latency and energy. Such a model-free algorithm permits to capture the unknown offloading cost variations under oblivious adversary, e.g., weighted-average randomized selection rule, biased cost estimation, e.g., implicit-exploration, and data privacy considerations, e.g., full-bandit feedback. 
	 
	\item The proposed algorithm facilitates to make an offloading decision in a manner adaptive to the volatile and time-varying resource necessitate and provision by appropriate adjustment on cumulative learning score in the selection rule. As such, the modified learning score considering the coming task demand and evolving circumstance, allows to select a suitable fog node, rather than a capable one, i.e., choosing a VFN better suited for the next task with time-varying size, rather than the one providing the lowest service cost for the current task.

	\item The proposed algorithm makes it possible to alleviate the uncertainty of the empirical cost estimates in assessment rule. The robust learning based on implicit exploration which controls the variance at the price of introducing some bias could guarantee near-optimal performance, rather than exploring the sub-optimal actions due to large variability attributable to unbiased estimation process.  
	
	\item The theoretical analysis about efficiency of the proposed algorithm is provided in terms of learning regret. It is proved that such a modified implicit exploration approach renders the reduction of variance and bias simultaneously, thereby achieving a better exploitation-exploration balance in an adversarial environment. Simulation results {in synthetic and real-world scenarios} verify its effectiveness and robustness. 
\end{itemize}

The rest of this paper is organized as follows: In Section II, the system model and problem formulation are presented. In Section III, the task offloading algorithm is then proposed. In Section IV, and the learning regret is analyzed. Simulation results are then provided in Section V, and finally comes the conclusions in Section VI.

\section{Related work}

This section presents related works in the area of VFC enabled task offloading, in terms of the potential scenarios of VFC and the task offloading algoritihms.



%
%
%
%
%
%
%

{ 
	\subsection{Task offloading scenarios}
    A variety of use cases have been identified as potential scenarios for VFC, i.e., efficient dissemination of real-time vehicle traffic and emergency information in cooperative driving of autonomous vehicles for road safety and intelligent firefighting for rescue safety \cite{primo}. In particular, emerging assisted driving applications, such as real-time situational awareness \cite{Hu2017}, lane changing and seethrough for passing \cite{Sabella2017}, and localization/mapping applications, such as HD map generation and road construction detection \cite{Zhu2018}, involve time-critical and computationally intensive tasks, i.e., on-road object recognition and scene understanding from images/videos, and have requirements for the validity period of task and reducing its consumed power. Safety-related services require low-latency responses, such as 10ms for cooperative collision avoidance, 25ms for vehicle platooning, and 500ms for collective environment perception \cite{3GPP2017}. The work in \cite{Zhu2018} investigated the feasibility and challenges of applying VFC for real-time analytics of high-resolution video streams, and proved the efficiency of VFC-based task offloading in terms of the latency, packet loss ratio and throughput. The work in \cite{Xiao2018} reduced the offloading latency considering the efficiency of power usage in fog computing by balancing the workload of fog nodes. Powerful computers are required for processing computationally complex tasks. Such computers also consume energy at a high rate, which affects vehicles' driving endurances if computers are powered by vehicles. The work in \cite{Lin2018} indicated that the driving distance reduced by 6\% due to the consumption of a computing engine equipped with 1 CPU, Intel Xeon E5-2630, and 3 GPUs, NVIDIA TitanX. 
%
%
%
%
%
%
%
%
%
     
     . 
    
}

 \begin{figure}[t]
	\centering \hspace{-1.5em}
	\includegraphics[width=0.34\textwidth]{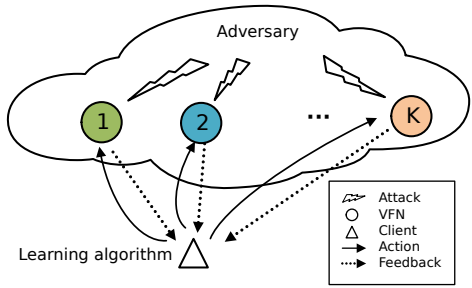}\vspace{-1em}
	\caption{Online learning in an adversarial envrionment.}\vspace{-1em}
	\label{LearningDiagram}
\end{figure}

   	\subsection{Task offloading algorithms}
   	Some efforts have been made to address the decision-making strategies for VFC-based task offloading. Specifically, the works \cite{Feng2017, Feng2018, sun2018, wu2018, sun2019, Liu2020, Zhu2019, Zhang2020} designed decentralized task offloading strategies where offloading decisions are made by the task generators independently. {The works in \cite{Feng2017, Feng2018} proposed task assignment algorithms for VFC enabled system, without centralized control, according to the collected information of adjacent vehicles. The proposed task caching and Ant Colony Optimization (ACO) based algorithm in \cite{Feng2017} attains efficient time complexity than brute-force approach, but may suffer from high complexity for a large number of vehicles and failure to adapt to a volatile environment. In \cite{Feng2018}, the task is processed in an online manner, but the proposed algorithm may suffer from high signaling overhead, i.e., heavily relying on frequent state information exchange, and thus failure to process subsequent tasks properly in case of outdated information provided by vehicles. To overcome such scalability issues, learning-based task offloading schemes have been considered in \cite{sun2018, wu2018, sun2019, Liu2020, Zhu2019, Zhang2020}.}
   	
   	The work in \cite{sun2018} proposed a learning-based task replication algorithm based on combinatorial MAB, where task replicas can be offloaded to multiple vehicles to be processed simultaneously. Some enhancements to this approach were achieved by adjusting the exploration weight according to the computation workload \cite{wu2018} and the appearance time \cite{sun2019} of fog nodes. The work in \cite{Liu2020} proposed a fluctuation-aware learning-based computation offloading algorithm based on MAB, where base stations are regarded as agents to learn the state of moving server. The work in \cite{Zhu2019} proposed an efficient online task offloading strategy to minimize the long-term cost of non-stationary fog-enabled networks. The work in \cite{Zhang2020} considers a mortal bandit formulation to address the time-varying set of VFNs for a given task generator, where the computation capacities of the edge nodes are used as contextual information in order to reduce the exploration space. However, all previous works assume that the task offloading performance experienced by an offloading service client is in a stochastic domain where some private information could be inferred by an attacker, and will be severely compromised by the non-stochastic task offloading strategies of other devices, i.e., the task offloading problem is adversarial and conventional upper confidence bound-based task offloading algorithms cannot be directly applied in an arbitrary dynamic environment.
   	
   	To solve the non-stochastic task offloading problem, adversarial MAB approach can be considered, where each strategy is assigned an arbitrary and unknown sequence of rewards, one for each time step, chosen from a bounded real interval. Especially, Exponential-weight algorithm for Exploration and Exploitation (Exp3) is a well-known learning algorithm for adversarial setting, and has been studied in resource provider selection problems \cite{Wang2018, Marcastel2019, Zhou2021}. Exp3-based online scheme has been proposed with the objective of optimizing the QoS, such as the throughput \cite{Wang2018}, energy consumption \cite{Marcastel2019} and latency \cite{Zhou2021}. However, the previous works fail to address mobility-induced volatile resource availability and resource demand in an adversarial environment at the same time.

\section{System model and problem formulation}
In this section, the system model and problem formulation are considered, applicable to offloading services.

\subsection{System model}
An offloading service client generates tasks, while a set of offloading service providers $k \in \mathcal{K} = \{1,...,K\}$ support the requested tasks with their own available computational resources. Any vehicles on the ground could become task offloading service clients or service providers. A service client $n$ can offload a task $t$ to any VFN $k$ within its communication range. Here $k \in \mathcal{K}^t \subseteq \mathcal{K}$ where $\mathcal{K}^t$ is the candidate VFN set varied due to their inherent mobility. VFNs available to a service client are discovered and selected by the client based on their topological states, {including moving direction and speed}\cite{Zhu2018}. For example, VFNs periodically broadcast single-hop messages including such state information following vehicular communication protocols, such as dedicated short-range communication (DSRC) or cellular vehicle-to-everything (C-V2X). {Each client forms a candidate VFN set by selecting from the accessible VFNs which follow the same driving direction as the client.} Due to inherent mobility, the candidate VFN set $\mathcal{K}^t$ varies. It is assumed that the client interacts with accessible VFNs continuously and updates the candidate VFN set $\mathcal{K}^t \neq \emptyset$\footnote{If $\mathcal{K}^t = \emptyset, \forall n, t$, the task $t$ may be processed locally or forwarded to a remote cloud server, which is left for the future work.} in real-time.

\subsubsection{Demand model}
In general, computing tasks can be divided into subtasks at different levels of granularity \cite{2016Akherfi}, and can be divided into atomic tasks\footnote{A task with a larger workload can be further partitioned into multiple ones.}. Multiple divisible subtasks can be executed in a parallel, serial or mixed manner. Some of these subtasks must be performed locally and some of them can be either performed locally or offloaded to the external computing resources. In this work, each atomic task is considered as a basic unit for offloading, i.e., offloaded to and processed by a fog node within one time period, and the operational timeline is discretized based on the atomic task unit, $t \in [t, t+1)$\cite{Zhang2020}. One can characterize an atomic task, $t$, by two parameters, the input size $q^t$ (bits/task) and its required computation resource defined as the number of CPU cycles $c^t$ (cycles/task). The resource demand can be estimated from measurements by applying the methods described in \cite{2018Neto}, and expressed as the multiplication of two parameters, the input size $q^t$ and the computational complexity $w^t$ representing the number of CPU cycles required for processing one bit of input data. The value of $w^t$ varies with applications, depending on the nature of performed applications\footnote{$w^t$ is approximated by a Gamma distribution in \cite{2010Miettinen}, e.g., face recognition requires 2339 (cycles/bit) and video transcoding requirement varies from 200-1200 (cycles/bit)}.


\subsubsection{Resource model}
The computational capability of a fog node $k\in \mathcal{K}^t$ is described by its maximum CPU frequency $F_k$ (cycles/second). One atomic task is offloaded as a whole to a single fog node who may execute tasks in parallel depending on its own resource allocation rules. To deal with multiple computation tasks simultaneously, a fog node dynamically adjusts its CPU frequency with dynamic frequency and voltage scaling (DFVS) technique. This work considers that the computing capability allocated to a fog node $k$, denoted by $f_{k}^t$ (cycles/second), is determined by the computing resource allocation policy, remains static for each task $t$ and in general is a non-increasing function of the total number of clients that offload to the same fog node $k$. It is assumed that each fog node employs equal fair resource scheduling over different tasks. The wireless medium of a fog node $k$ is shared by the clients that choose to offload to the fog node $k$. The achievable uplink and downlink transmission rates between a client and fog node $k$ are determined by the physical characteristics of the wireless medium, such as distance, fading gain, bandwidth, and interference.

\subsubsection{Cost model}
Performing task offloading incurs transmission and computation costs\footnote{The term cost is often interchangeable with the loss in this work}. Two kinds of cost can be considered, the offloading service latency $L$ and the related energy consumption $E$. Specifically, the latency for offloading includes the time for uploading the input to a fog node $k$, and the execution time at the fog node, downloading the result to the service client. It is assumed that the feedback size is small enough that the downlink transmission latency can be safely ignored. Thus, the latency of completing a task is expressed as 
\begin{eqnarray} 
D_k^t &=& \frac{q^t}{r_{k}^t} + \frac{q^t w^t}{f_{k}^t} \nonumber
\end{eqnarray}
where $r_{k}^t  = B\log\left[1+ \frac{P g_{k}}{N + I_k}\right]$ is the link rate for transmitting input data $t$ from a client to a fog node $k$, $B$ is channel bandwidth, $P$ denotes transmission powers of a client, $g_{k}$ is the uplink channel gains between a client and a fog node $k$, $N$ is the noise power, and $I_k$ denotes interference measured at the fog node. Given the orthogonal channel allocation \cite{Kenney2011}, the co-channel interference can be avoided. Furthermore, the cross-channel interference can be ignored according to the experimental results in \cite{Rai2007}. The channel gains are static during the uploading process of each computation task and downloading processes of the computation result. The energy consumption of completing a task is expressed as
\begin{eqnarray}
E_k^t &=& \frac{P\cdot q^t}{r_{k}^t} + \frac{P_k^t q^t w^t}{f_{k}^t} \nonumber
\end{eqnarray}
where $P_k^t = \rho (f_k^t)^3$ is the computing power with effective switched capacitance related to the chip architecture\footnote{Local execution does not cause transmission cost, but computation cost in terms of latency and energy consumption. In this work, a task is assumed to be for being transmitted to a fog node.}\cite{Wang2016}. To take into account two types of costs for task $t$ to a VFN $k$, we define the cost function as the weighted sum over the latency and energy consumption, 
\begin{eqnarray} \label{general_utility}
U(t,k) &=& \xi D_k^t + (1-\xi)E_k^t
\end{eqnarray}
where $\xi$ denotes the weighting parameter of latency~\cite{Qin2020}.

\subsection{Problem formulation}
{ 
We define the unit cost of task offloading as the overall cost of offloading the processing of one bit of input data for task $t$ to a VFN $k$, 
\begin{eqnarray} \label{general_utility2}
l(t,k) &=&  {U(t,k)}/{q^t} = \xi D_{k,o}^t + (1-\xi)  E_{k,o}^t
\end{eqnarray}
where $D_{k,o}^t = D_k^t/q^t$ and $E_{k,o}^t = E_k^t/q^t$ denote the per-bit latency cost and per-bit energy cost, reflecting the service capability of each candidate VFN $k$. We aim at minimizing the average unit cost of task offloading by optimizing the fog node selection, done by each client, for each task (up to a finite $T$ tasks) in each round, $k_t$. If all state information related to the per-bit cost are exactly known by the task requester before offloading each task, the optimization problem can be expressed as follows: $k_t = \min_{k\in \mathcal{K}_t} l(t,k)$ where $k_t$ is the optimization variable representing the index of fog node selected for task $t$, $k_t \in \mathcal{K}_t$. 
 
In fact, the state information of fog nodes in heterogeneous and dynamic networks is hard-to-predict and exchanging the information among the clients and fog nodes causes high signaling overhead. Thus, the clients may lack the state information of fog nodes and could not make accurate predictions about which fog node would provide the optimal offloading service for each task. To overcome this, one may utilize learning-and-adapting-based offloading scheme where a client observes and learns the costs of each task offloaded to the candidate fog nodes and makes an offloading decision based on the historical cost observations without exact knowledge about the current state information. For this, we aim to design a learning-based algorithm minimizing the expectation of the unit offloading cost, formulated as
\begin{eqnarray} \label{eq:P0}
	\mathcal{P}: \min_{k_1, k_2, \cdots, k_T} \mathbb{E}\left[\sum_{t=1}^T l(t,k_t)\right]
\end{eqnarray}
where $\mathbb{E}\left[\cdot\right]$ is the expectation operator, $l(k,k_t)$ is a sequence of unit cost for the $t$-th task in the set of tasks $\mathcal{T}$, and $T = |\mathcal{T}| \in \mathbb{N}^+$ is the number of tasks. 
}

\section{Online learning-based task offloading}
In this section, a learning-based task offloading algorithm is developed based on MAB, which enables a client to learn the offloading cost of candidate fog nodes and optimizes the expected task offloading cost. The problem (\ref{eq:P0}) requires online sequential decision making whose nature enables to design a lightweight algorithm but suffers from uncertainty associated with the lack of knowledge about the properties and conditions of the phenomena underlying the behaviour of the systems.

\subsection{Learning under uncertainty: an adversarial approach}
Consider a general framework of online learning where a task client selects one fog node, $k$ from a finite set $\mathcal{K}^t$ based on an a priori unknown payoff function. The previously offloaded tasks allow an empirical mean as an estimate of the expectation, but if there are not enough observations, this guess may not be accurate. In order to get more information about one specific fog node, the client needs to offload more tasks to that node even though it may not be the empirically best fog node to offload. However, the empirically best node is preferred for the sake of instantaneous benefits in online decision-making. Therefore, there exists a tradeoff between exploiting the empirically best node for instantaneous rewards and exploring other nodes for potential benefits. Also, note that learning under uncertainty relies on feedback in general. Thus, quality\footnote{While many works take assumption requiring the noise to have a well-conditioned, stochastic component, i.e., independent, identically distributed Gaussian process noise, imperfect feedback referring to the inaccuracy of the observed utilities in revealing the quality of the selected actions is assumed to be null in adversarial regime due to its arbitrary property. One may further consider adversarial noise sequence which is left for the future work.} of the feedback in terms of completeness has significant implications in assessment rule. Incomplete feedback stands in contrast to full-information feedback where utilities of all actions a client could have taken are observed in each stage. Incompleteness can be spatial/temporal across the action space/stages. When a client sends a task to a fog node, there is no way to know how other fog nodes would have performed on the same task. Moreover, local visibility of loss makes decision-making challenging. A commonly studied model is the so-called bandit feedback, where only the utility of the chosen action is revealed. The term bandit feedback has its roots in the classical online learning problem to play a multi-armed slot machine known as a bandit. A MAB problem is specified by a set of arms (actions or available VFNs) $\mathcal{K}^t$ and a sequence of cost $l_{k}^t, t\in \mathcal{T}$. For each task, a client selects an arm and receives the cost from the selected arm, not from other arms.

The objective of a client is to minimize the long-term cost as shown in equation (\ref{eq:P0}), while managing exploration-exploitation trade-off in bandit setting. Each arm pulled by the client generates cost in an adversarial fashion. An adversary is changing the future cost for arms, and the distribution of cost for each arm would change over time, which is not inherently probabilistic and does not include stochastic averaging in contrast to stochastic MAB case. In this sense, non-stochastic formulation of MAB is more appropriate to evaluate the most promising strategy in an arbitrarily changing environment where there could even exist an oblivious adversary, e.g., jamming attack. Also, existing stochastic MAB may characterize the exploration bonus in determined selection rule with a padding function addressing the variations of additional informative data such as the history of playing up to the current round, which would result in better performance. However, incorporating such information into an adversarial setting is challenging due to randomized arm selection rule, and the payoff generated in an adversarial fashion under information limited environment weakens the robustness and smoothness of the estimation process in its subsequent assessment rule.

\subsection{Exploration in selection rule}\label{sectionIII-B}
In an adversarial MAB problem, randomized policy is needed due to the possibility that a client using deterministic policy or stochastic one such as Upper Confidence Bound (UCB)-style exploration \cite{Bubeck2012} may be easily fooled by adversaries. Thus, instead of choosing an arm $k' \in \mathcal{K}^t$ directly, the client $n$ selects a probability distribution $\Lambda^t = [p_{k}^t]_{k\in \mathcal{K}}\in [0,1]^{|\mathcal{K}^t|}: \sum_{k\in \mathcal{K}^t} p_{k}^t = 1$ over the available arms for task $t$. The resulting probability vector $\Lambda^t$ is called a mixed strategy for the mixed strategy space of a client who draws an arm according to this distribution, $k' \sim p$. The selected probability distribution is proportional to its loss, weighted appropriately. The idea is to give more weight to actions that performed well in the past. One may employ weighted-average randomized strategy with potentials\footnote{Polynomial and exponential potentials can be used \cite[Section 6]{cesa2006}.} to achieve a cumulative cost (almost) as small as that of the best action \cite[Section 6]{cesa2006}. An arm $k$ is assigned with the selected probability for task $t$, $p_{k}^t$ which is proportional to weighted accumulated cost caused by that arm in the past, $p_{k}^t = \frac{\mathcal{W}_{k}^t}{\sum_{k} \mathcal{W}_{k}^t}$. The parameter, $\mathcal{W}_{k}^t$, is a weight of each arm $k$ maintained by the client, representing the confidence that the arm is a good choice.

In a bandit setting, rather than concerning about how to get the estimated cost of an arm which was not pulled, one seeks to investigate how such information can be used when it becomes available. To that end, the score (penalty) based learning process is considered as follows: The service capability of a fog node can be represented by the score parameter, the cumulative estimated bit-per cost up to $s-1$, $\mathcal{\hat{L}}_{k}^{s-1} = \sum_{t=1}^{s-1} \eta_t \hat{l}_{k}^t$ where $ \hat{l}_{k}^t$ is the estimate of loss from the arm $k$ for task $t$ and $\eta_t\in (0,1]$ is the learning rate. If all of arms newly appear in round $t = 1$, their scores are initialized with zero, $\mathcal{\hat{L}}_k^{0} = 0,~\forall k\in \mathcal{K}^{1}$ and thus the resulting probability follows a uniform distribution initially, $p_k^1 \sim {1}/{|\mathcal{K}^{1}|}$. In each round $s$, a task requester chooses an action $k'$ based on the resulting probability $\Lambda^s$ and updates the estimate of loss $\hat{l}_{k}^s$ based on the selected arm. The resulting probability $\Lambda^s$ is determined based on the scores $\mathcal{\hat{L}}_k^{s-1}$.

Essentially, one would leverage past experiences to gain the intuition on what is the best value to use. Considering the exponential potential function with the score, the weighting parameter can be expressed as $\mathcal{W}_{k}^s = e^{-\mathcal{\hat{L}}_k^{s-1}}$, and the resulting probability vector $\Lambda^s$ is expressed as $p_{k}^s = \frac{ e^{{-\mathcal{\hat{L}}_k^{s-1} }}}{ \sum_{m} e^{{- {\mathcal{\hat{L}}_m^{s-1}}}}}, \forall k\in \mathcal{K}^s$ in round $s$. Note that such importance-weighted mechanism assigns exponentially higher probability to strategy with lower cumulative scores up to $s-1$ due to the relation $\frac{\partial}{\partial x} \left( \frac{e^{-x}}{X + e^{-x}}\right) =  \frac{-e^{-x}X  }{(X+ e^{-x})^2}<0$ where $x = \mathcal{\hat{L}}_k^{s-1}$ and $X = \sum_{m\neq k} e^{{- {\mathcal{\hat{L}}_m^{s-1}}}} $. These scores reinforce the success of each strategy measured by the estimated offloading cost $\hat{l}_{k}^{s-1}$, so a client would rely on the strategy with the lowest score.

Appropriate selection rule could achieve a balance between exploitation and exploration, i.e., exploiting known resources with certainties and exploring for new possibilities, by differentially choosing among actions, favoring those with lower cumulative scores perceived to be more attractive. While the exploration and exploitation trade-off conventionally depends on the service provider's state i.e., candidate fog node' capability, its balance can be improved by considering additional information on the service requesters' necessity and providers' activities. However, in an adversarial setting, it is nontrivial to improve the performance with such additional information. Non-stochastic property nullifies the statistics of the historical data on the service providers' activities, i.e., the numbers of computational tasks that an arm has served and for which it has been connected to the requester since its initial connection \cite{sun2019}, are void especially in a volatile dynamic environment. Also, a padding function in the UCB based selection rule allows to characterize an exploration bonus addressing such informative data, while it is not straightforward to do that in an adversarial setting due to its randomized policy.

This work aims at incorporating the observation on the resource provider's volatility and resource requester's task size into the selection rule in an adversarial setting to achieve a better balance between exploration and exploitation. To do so, in the following, the dynamic resource supply and demand-based exploration bonus is augmented in the score $\mathcal{\hat{L}}_{k}^{s-1}$ toward fair and suitable fog node selection.

\subsubsection{Dynamic resource supply}
If an arm $\bar{k}$ newly appears in round $\tau$, $\mathcal{K}^{\tau} = \mathcal{K}^{\tau-1}\cup \bar{k}$, as the previous candidate set of fog nodes did, i) all arms including the new arm could be reset, $\mathcal{\hat{L}}_{{k}}^{\tau-1} = 0$, named full reset, or ii) only the new arm's score could be initialized with zero, $\mathcal{\hat{L}}_{\bar{k}}^{\tau-1} = 0$, named partial reset. However, such a resetting mechanism may invalidate the score based learning benefit in the rapidly changing environment. The bandit may take a long time to collect enough samples for those arms to correct their null scores again. Also, such incomparable scores due to the partial nullification may fail to fairly explore all of the available arms to identify the best arm within a total number of tasks, $T$. For instance, if the existing scores of the arms $\mathcal{\hat{L}}_k^{\tau-1}, \forall {k} \in \mathcal{K}^{\tau-1}$ are as high as those in making the corresponding resulting probabilities too low, the newly appeared arm $\bar{k}$ will be dominant $p_{\bar{k}}^{\tau} \gg p_k^{\tau}, \forall {k} \in \mathcal{K}^{\tau-1}$ and thus repeatedly be selected for all eligible rounds $\tau' > \tau$ until the arm's score goes up enough to make more accurate estimation comparative to other arms $s>\tau'$. In other words, the old arms may sacrifice their opportunities, regardless of their accumulated experience, to learn the dynamic task offloading environment. Such an unfair selection rule from the perspective of old arms could be amended by setting the initial score of an appearing arm with the already existing one from oneself or others $\mathcal{\hat{L}}_{\bar{k}}^{\tau-1} = \mathcal{\hat{L}}_{\bar{k}}^{\tau-1} + \beta_{\bar{k}}$ where $\beta_{\bar{k}} = \max(\min(\mathcal{\hat{L}}_{m}^{\tau-1}),\mathcal{\hat{L}}_{\bar{k}}^{\tau-1}), m\in \underline{\mathcal{K}}^{\tau-1}>0$ and $\underline{\mathcal{K}}^{\tau-1}$ is the set of the old arms in round $\tau-1$.

\subsubsection{Dynamic resource demand}
Note that while the objective in equation (\ref{eq:P0}) is to optimize the expected bit cost of offloading the task to a fog node $k$ for task $t$, what actually needs to be learned is the potential capability of each candidate fog node and its projected suitability for the upcoming task under an adversarial framework. The service suitability of a fog node can be assessed by the normalized total delay of offloading the next task $\mathcal{\hat{L}}_{k}^{t-1} q^t$ which would be further additive to the service capability to build refined weights for better arm selection and thus improved quality of service, e.g., cost per task. Such joint consideration of both the normalized offloading delay per bit and per task $\mathcal{\hat{L}}_{k}^{t-1}(1+ q^t)$ may take some coordination in terms of input data size-dependent exploration-exploitation trade-off. For the feature scaling, the normalized size of the upcoming task $q^t$ is used as a weight factor ${\delta}^t = 1+({q^t}-q_{min})/({q_{max}-q_{min}})$ where $q_{max}$ and $q_{min}$ are the upper and lower thresholds of the input data size, respectively, on the offloading delay in decision-making algorithm, i.e., $\mathcal{W}_{k}^t = e^{-\mathcal{\hat{L}}_{k}^{t-1}{\delta^t}}$. This approach turns out to be analogy to the Boltzmann (or softmax) exploration~\cite{Barto1991}, which creates a graded function of estimated value with the maximum inverse temperature parameter equal to 2~\cite{Tijsma2016}. The higher values of $\delta^t \rightarrow \infty$ will lead to a fully greedy strategy, while the lower values $\delta^t \rightarrow 1$ will move the selection strategy more towards offloading service capability-based one.

\subsection{Exploration in assessment rule}
According to the selection rule above, one client selects a suitable fog node for the upcoming task, offloads it to the selected node, and receives real-valued payoffs, i.e., offloading service cost per bit, and then uses its own assessment rule to independently convert the realized payoff into the learning-weighted estimate of the payoff additive to the previous score representing the fog node's estimated capability.

\subsubsection{Iteration-varying learning} 
Learning rate is a parameter controlling how much the weights of the current estimated payoff is taken into account for the upcoming cumulative score, which determines the importance of the estimated payoff at each time in term of contribution to the cumulative score. Conventionally, the learning factor $\eta_t$ is predefined as a empirically constant or variable depending on the horizon $T$, which requires advance knowledge of the horizon and weakens the learning ability of algorithm. Note that achieving the perfect knowledge of $T$ is usually not feasible in practice. While one could use a standard doubling trick~\cite{Auer2002} to overcome this difficulty, we choose to take a different path to circumvent this issue, and propose to tune its learning rate iteration-dependently $\eta = \eta_t, \forall t$ and other parameters solely based on observation. Thus, from technical perspective, the task requester should take positive actions to explore unfamiliar environment and learn the loss statistics $\hat{l}_{k}^t$ of all strategies in the initial stage. As learning iteration goes on, the client may want to exploit observations obtained so far to identify the best strategy without engaging others too often.

However, it is nontrivial to select a proper $\eta_t$ which should be large enough to avoid selecting a bad arm too many times, while small enough to limit the transient effect. One way is to encourage an algorithm to explore less over round, decreasing learning factor with round; the more distant the past, the more its learning factor. When the learning rate is large, $p_{k}^t$ becomes more uniform, and the algorithm explores more frequently. For a lower learning rate, $p_{k}^t$ concentrates on the arm with the lowest estimated cost and the resulting algorithm exploits aggressively. Furthermore, if the exploration-exploitation levels change too fast, it would be too short to obtain the inflection point from exploration to exploitation. For this matter, one may further consider varying the learning factor with the number of candidate sets; the larger the number of arms is, the more slowly the learning factor decreases.

\subsubsection{Robust learning} 
Learning algorithms are based on a model of reality, and their performance depends on the degree of agreement on their assumed model with reality. The robustness of an algorithm is its sensitivity to discrepancies between the assumed model $\hat{l}_k^t$ and the reality ${ {l}_k^t}$, which is essentially determined by how the related assessment rule is set.

The loss from an arm  $k \neq k'$ could not be observed due to incomplete feedback in the bandit problem. This motivates us to use unbiased estimate that the client observes, enabled by i) using the loss ${l}_{k}^t$ if one observes it and $0$ otherwise, $\hat{l}_{k}^t = {l}_{k}^t\cdot \mathbbm{1}_{k = k'}$, and ii) correcting the bias from dividing it by the probability of selecting the arm, $\hat{l}_{k}^t = {{l}_{k}^t}/{p_{k}^t}\cdot \mathbbm{1}_{k = k'}$, thereby maintaining the expectation property and making arms that have not been pulled yet optimistic and being likely to be explored. However, the unbiased estimate causes large fluctuation in the loss due to inverse-proportion to $p_{k}^t$. One idea is to avoid $p_k^t$ being too small. The first thing that comes to mind is to mix $p_{k}^t$ with the uniform distribution. This is an explicit way of forcing exploration, which after further modification can be made to work. The idea to reduce the variance of importance-weighted estimators has been applied in various forms \cite{Uchibe2004, Wawrzynski2007, Bottou2013}, but all of these works are based on truncating the estimators, which makes the resulting estimator less smooth.

This work takes the similar approach for a simpler and empirically superior algorithm. They key idea is to change the cost estimates to control the variance at the price of extra bias. To achieve this, we consider Exp3 algorithm endowed with implicit exploration (IX)-style cost estimates \cite{Neu2015}. After each action, the cost is first calculated as $\hat{l}_{k}^t  =  {{l}_{k}^t}/({p_{k}^t + \gamma_t }) \cdot \mathbbm{1}_{k = k'}^t$, which is a biased estimator due to $\mathbb{E}[\hat{l}_{k}^t] = \sum_{k } p_{k}^t \hat{l}_{k}^t  = {l}_{k}^t -  \frac{\gamma_t \cdot l_{k'}^t}{p_{k'}^t+   \gamma_t} \leq l_{k}^t$ where $p_{k}^t$ is the probability, percentage of weight, that arm $k$ will be chosen for task $t$. The implicit exploration parameter $\gamma_t \in (0,1]$ makes $p_{k}^t$ smooth so that actions with large losses for which classical recipe in exponential weights algorithm scheme, would assign negligible probability, are still chosen occasionally, and thus the estimator is allowed to guarantee reliable performance in rapidly changing, adversarial environments.

\subsection{Proposed algorithm} 
In this work, taking into account the above mentioned motivations, an algorithm for adaptive learning-based task offloading is proposed to solve the offloading decision problem where a client decides for each task to which fog computing node to offload it. The proposed algorithm makes use of two exploration processes. One is modified Boltzmann distribution based exploration that supports time-varying resource supply and demand dependent offloading, emphasizing feasibility and fairness in fog node selection. The other is an implicit exploration based on biased loss estimation to alleviate the uncertainty of the importance-weighted estimator.

\begin{algorithm}[t]\small 
	\caption{MIX-AALTO: Modified Implicit Exploration based Algorithm for Adaptive Learning-based Task Offloading}  \label{algorithm1}
	\begin{algorithmic}[1]  
		\State Input: sequences $\eta_t>0$, $\gamma_t>0$ , $\mathcal{K}' = \emptyset$
		\For{$t\in \mathcal{T}$}   
		\State Set $\eta \leftarrow \eta_t, \gamma \leftarrow \gamma_t$, $q \leftarrow q^t, {\mathcal{K}} \leftarrow \mathcal{K}^t, {\mathcal{K}'} \leftarrow {\mathcal{K}'} \cup \mathcal{K} $ 
		
		\State Set ${{L}}_{k} \leftarrow 0$, $\beta_k \leftarrow 0, k\in \mathcal{K}$  \Comment{Dynamic supply}
 	
		\For{Any $k\in \bar{\mathcal{K}}$}  
		

		\If{$k\in \bar{\mathcal{K}}\backslash (\bar{\mathcal{K}} \cap {\mathcal{K}}')$}
	\State	Update $\beta_k \leftarrow \min(\mathcal{\hat{L}}_{m}), m\in \underline{\mathcal{K}}$
		\EndIf	
		
		\If{$k\in \bar{\mathcal{K}} \cap {\mathcal{K}}'$}
		\State	Update $\beta_k \leftarrow \max(\min(\mathcal{\hat{L}}_{m}),\mathcal{\hat{L}}_{{k}}), m\in \underline{\mathcal{K}}$
		\EndIf	
		
		\EndFor	
 
\State Update ${{L}}_{k} \leftarrow \mathcal{\hat{L}}_{k}, k\in \underline{\mathcal{K}}$   
\State Update $\delta \leftarrow q$		\Comment{Dynamic demand}

	    \State Update $\mathcal{W}_{k} \leftarrow  \delta \cdot (L_{k}+ \beta_k)$ \Comment{Selection rule}
		\State Set $p \leftarrow \left[\frac{\exp(-\mathcal{W}_{k})}{\sum_{m} \exp(-  \mathcal{W}_{k})}\right]_{k\in \mathcal{K} }$ 
		\State Select action $k' \sim p$
		\State Receive the cost ${l}_{k'} \leftarrow U_{k'}$ \Comment{Assessment rule}
		\State Compute $\hat{l}_{k}  \leftarrow \left[ \frac{{l}_{k} \cdot \mathbbm{1}_{k = k'}}{p_{k} +   \gamma}\right]_{k\in \mathcal{K} }$
		\State Update scores: $\mathcal{\hat{L}}_{k} \leftarrow \hat{\mathcal{L}}_{k}  + \eta  \hat{l}_k, \forall k$
		\EndFor 
	\end{algorithmic}
\end{algorithm}
   
 In Algorithm~\ref{algorithm1}, the vanishing learning factor $\eta_t$, exploration factor $\gamma_t$, and the set of previously used fog nodes $\mathcal{K}'$ which is here assumed to be empty initially, are considered as the input parameters (Line 1). And then the iteration dependent parameters, $\eta_t$, $\gamma_t$, $\mathcal{K}^t$ and $q^t$, are set. {Upon generating each task from the application the information on the input data size $q^t$ is known by the task requester. Also, the up-to-date information on a set of candidate VFNs $\mathcal{K}^t$ from the neighbor discovery process is available and $\mathcal{K}'$ is updated (Line 3).} Afteward, the algorithm is structured in three parts: i) exploration bonus adjustment where two dynamic factors in terms of resource supply and demand, $\beta_k$ and $\delta$, are updated, which would be used to tune the weighting parameter, $w_k^t$ (Lines 4-14), ii) selection rule domain where the selected probability is proportional to the cumulative score tuned by considering the resource demand and supply aspects for suitable and fair selections via modified exploration bonus (Lines 15-17), iii) assessment rule domain where the utility function defined in Eq. (\ref{eq:P0}) is used to evaluate the service capability of each fog node, by observing the empirical offloading cost and converting it to the estimated cost via implicit exploration factor, and then the cumulative learning score is updated (Lines 18-20).
 
 { 
 \subsubsection{Adaptivity}
Adaptivity is an essential property that has steadily gained importance for solving the offloading decision problem, particularly for dynamic environments. For the adaptation to dynamic resource supply and demand, a cumulative learning score is fine-tuned with parameters $\beta_k$ and $\delta$ such that the available arms are fairly and suitably explored. While the parameter $\delta$, identical for all candidate VFNs, plays a role in modifying the explore-exploit behavior for demand dependent suitability, the parameter $\beta_k$, possibly different for different VFNs, plays a role in reducing the large disparity between the cumulative learning scores of different VFNs, thereby avoiding unfair selection.
Prior to performing the offloading decision for each task, a neighboring VFN set is discovered within its communication range \cite{Zhu2018} and those in the same moving direction are considered as candidate VFNs \cite{sun2019}. Due to inherent mobility-induced time-varying features, some fog nodes physically leave a candidate set temporarily but return into the set within a finite number of time periods, called volatile occurrence of the potential candidate fog nodes. For the arm {$k\in \bar{\mathcal{K}}^t\backslash (\bar{\mathcal{K}}^t \cap {\mathcal{K}}')$} newly appearing in the candidate set for task $t$, the corresponding score $\beta_k$ is set to be the minimum score of the other existing available arms (Line 7). If the arm {$k\in (\bar{\mathcal{K}}^t \cap {\mathcal{K}}')$} which has ever been connected to the task generator becomes available again after not so long time, the previously used score is re-utilized so that it leverages to its own recently estimated computation capability, rather than other arms (Line 10). Such volatile resource supply based score assignment allows discovered candidate arms to be fairly explored with the highly capable existing arm or their own knowledge, and the algorithm to adapt to the change. Regarding the adaptation to resource demand, time-varying demand dependent offloading decision is enabled by joint consideration of both the normalized offloading bit-per cost and per-task cost in score, which results in a more suitable fog node selection, i.e., with a larger input size, more exploitation would be executed with firmed belief for a more suitable selection (Line 14). While such adaptivity to the dynamic resource supply and demand is treated in a modified form of exploration as exploration bonus in selection rule for fairness and suitability, the implicit exploration is considered to enable reliable cost estimation in assessment rule.

	\subsubsection{Scalability} 
	To cope with the heterogeneity in resource capacities and adaptivity in a dynamic environment, the proposed algorithm is to keep high scalability taking into account 
	i) computational complexity, e.g., time complexity,
	ii) communication overhead given by its implementation, i.e., how many times a decision-maker needs to communicate with available fog nodes, and 
	iii) accessible information, i.e., what type of information a decision-maker needs before making decisions. That is, the key properties of scalability are low complexity, low communication overhead, and reduced need for information.
	
	\begin{remark}
		(Low computational complexity) Calculating the modified scores of all candidate VFNs has a complexity of $\mathcal{O}(|\mathcal{K}|)$, and updating the estimated cost and cumulative score has a complexity of $\mathcal{O}(1)$. Thus, the proposed algorithm 
		has polynomical-time efficiency, i.e., $\mathcal{O}(|\mathcal{K}|  |\mathcal{T}|)$, which is lower than $\mathcal{O}(|\mathcal{K}| |\mathcal{T}|^2)$ for the ant colony optimization. 
	\end{remark}

	\begin{remark}
		(Low communication overhead) The proposed algorithm allows the task generator to learn the states such as allocated CPU frequency of each fog node, instead of obtaining them from physical signal messages, which can save $|\mathcal{K}^t|$ signaling messages for the state of the $\mathcal{K}^t$ candidate fog nodes.
	\end{remark}

	\begin{remark}
		(Low information demand) Instead of collecting all information via local observation and full feedback from available fog nodes, the proposed algorithm enables the task generator to make decisions based on a fully causal information oracle via local observation and bandit feedback.
	\end{remark}
	
}

\section{Learning efficiency of proposed algorithm}
This section characterizes the performance of the online learning algorithm. Naturally, exploring an uncertain world with a specific goal always has some regret. As a performance criterion, the considered assessment rule employs some notion of the learning regret which tries to capture the degree of cumulative dissatisfaction of a task generating client in presence of dynamic resource supply and demand.

\subsection{Regret}
Concretely, the regret of an algorithm is defined as its cumulative loss minus the cumulative loss of the best strategy in the pool, i.e., available candidate set. To address non-stationary environment where there is no single fixed point that does well overall, we use the regret with respect to an interval $\mathcal{T}_i = [\tau_{i},\tau_{i+1}-1] \subseteq \mathcal{T}  =[1, T]~\forall i\in \mathcal{I}$ where $\tau_{i+1}-1$ is the maximum of the rounds maintaining a network structure unchanged, i.e., available fog nodes are identical during an interval $\mathcal{T}_i$. The significance of no-regret learning depends on the adopted benchmark policy which the learning algorithm is measured against.

An oracle benchmark to $\mathcal{P}$ in equation~(\ref{eq:P0}), the optimal solution to the minimization problem during each interval, $t \in \mathcal{T}_i$ is given by $k^{*} \in \arg \min_{k} [ \bar{l}_{k}[i]]~\forall i$ where $\bar{l}_{k}[i]$ is the expectation of $l_k^t$, $E[l_k^t]_{t}$, for the interval $i$, which is unknown beforehand in practice. Given the oracle benchmark, the learning regret which measures how much the client regrets choosing his pulled action-sequence over the one with the optimal policy, can be expressed as 
\begin{eqnarray} \label{eq:P11}
	R^T =  E\left[L_{k'}^T  - L_{k^*}^T\right]
\end{eqnarray}
where $L_{k'}^T = \sum_{t\in\mathcal{T}} l_{k'}^t$
and $L_{k^*}^T =\sum_{t\in\mathcal{T}}l_{k^*}^t$  correspond to the sequences of cumulative losses incurred by the Algorithm \ref{algorithm1} and adopted oracle, respectively.

The regret upper bound of the proposed algorithm is analyzed, desirable to stay small in mean and concentrated well around the mean, so-called high-probability $1-\nu$ bounds. Such targeted properties guaranteed for each interval would be also valid for multiple intervals. Thus without loss of generality one may focus on an interval of the algorithm and omit the symbol index $i$, e.g., $\mathcal{T}_i = \mathcal{T}$. Such probability-based measurement value can be quantified as a concentration of measure inequality based on the Cramer-Chernoff method. 
The quantities of interest here are the variance and bias{\color{blue}, both of which would be used as bounded components of regret.}

{We show that the variance of the sum of a sequence of random variables cannot be much higher than the sum of their expectations conditioned on the past, following from \cite{Neu2015} using a martingale sequence and a Markov's inequality.}
\begin{lemma} \label{bounded_var} (Variance)
	With probability at least $1-\nu$, the following inequality $\sum_{k\in \mathcal{K}} (\hat{L}_k^T -L_k^T) = \sum_{k\in \mathcal{K}}  \sum_{t\in\mathcal{T}}  (\hat{l}_k^t - {l}_k^t)    \leq  {\ln(|\mathcal{K}|/\nu)}/{2\gamma_T}$ and $ \hat{L}_k^T -L_k^T =  \sum_{t\in\mathcal{T}}  (\hat{l}_k^t - {l}_k^t)   \leq  {\ln(|\mathcal{K}|/\nu)}/{2\gamma_T}$ for any fixed $k$. 
	\begin{proof} 
 		i) Define ${\lambda}^t = \sum_{k\in\mathcal{K}}  \alpha_t {l}_k^t$ and $\tilde{\lambda}^t = \sum_{k\in\mathcal{K}} \alpha_t   \tilde{l}_k^t $ where $\tilde{l}_k^t = \frac{{l}_{k}^t}{p_{k}^t+\gamma_t}$. Let $\gamma_t$ be a fixed non-increasing sequence and $\alpha_t$ be non-negative $\mathcal{F}^{t-1}$ measurable random variables. According to \cite{Neu2015}, if $\alpha_t\leq 2\gamma_t$  for all $t$, one gets $E[Z^t | \mathcal{F}^{t-1}]\leq Z^{t-1}$ where $Z^t = \exp(\sum_{\tau=1}^t (\tilde{\lambda}^{\tau} - {\lambda}^{\tau}))$ is super-martingale relative to $\mathcal{F}^{t-1}$ and $Z^0 = 1$. 
		ii) With Markov's inequality, one leads to $\mathcal{P}\left[\sum_{t\in\mathcal{T}}  \sum_{k\in\mathcal{K}} \alpha_t (\tilde{l}_{k}^t - {l}_{k}^t) > \epsilon\right] \leq   e^{-\epsilon} = \nu$ for any $\epsilon >0$, where $v$ is the probability that the bound is not satisfied. With the respective complement (guaranteed) probability $1- v$ and bandit feedback, one gets $\sum_{t\in\mathcal{T}}  \sum_{k\in\mathcal{K}}\alpha_t  (\hat{l}_k^t - {l}_k^t) \leq \ln(|\mathcal{K}|/\nu)$ and similarly, $\sum_{t\in\mathcal{T}} \alpha_t  (\hat{l}_k^t - {l}_k^t) \leq \ln(|\mathcal{K}|/\nu)$  for any fixed $k$.
		Thus, we obtain $\sum_{k\in\mathcal{K}} \sum_{t\in\mathcal{T}}   \gamma_t(\hat{l}_k^t - {l}_k^t) \leq \ln(|\mathcal{K}|/\nu)/2$ and $\sum_{k\in\mathcal{K}} \sum_{t\in\mathcal{T}}   (\hat{l}_k^t - {l}_k^t) \leq \frac{\ln(|\mathcal{K}|/\nu)}{2\gamma_T}$, and also $\sum_{t\in\mathcal{T}} \gamma_t (\hat{l}_k^t - {l}_k^t)   \leq     {\log(|\mathcal{K}|/{\nu})}/{2}$ and $\sum_{t\in\mathcal{T}}   (\hat{l}_k^t - {l}_k^t)   \leq     \frac{\ln(|\mathcal{K}|/\nu)}{2\gamma_T}$ for any fixed $k$.
	\end{proof}
\end{lemma}

Unlike a typical unbiased estimator, the cost estimator with implicit exploration parameter in the assessment rule incurs the bias which is the difference between the realized cost and the biased estimator's expected cost.

\begin{lemma}\label{bias}
	(Bias) With probability at least $1-\nu$, the bound on the bias is ${L}_{k'}^T -  \tilde{L}_{}^T  =  \sum_{k\in\mathcal{K}} \sum_{t\in \mathcal{T}}  {\gamma_t  \hat{l}_{k}^t } \leq  \sum_{k\in\mathcal{K}} \sum_{t\in \mathcal{T}} \gamma_t {l}_k^t +  {\log({|\mathcal{K}|}/{\nu})}/{2}$.
	\begin{proof}
	One gets the following: $\sum_{t\in \mathcal{T}} ( {l}_{k'}^t - E_{k\in\mathcal{K}}[\hat{l}_k^t]) = \sum_{t\in \mathcal{T}} ( {l}_{k'}^t - \sum_{k\in\mathcal{K}} p_k^t \hat{l}_k^t) = \sum_{t\in \mathcal{T}} ( {l}_{k'}^t - \sum_{k\in\mathcal{K}} {l}_k^t \mathbbm{1}_{\{k = k'\}}  +  \sum_{k\in\mathcal{K}} \mathbbm{1}_{\{k = k'\}} \frac{\gamma_t {l}_k^t}{p_k^t + \gamma_t})  =  \sum_{k\in\mathcal{K}} \sum_{t\in \mathcal{T}}   {\gamma_t  \hat{l}_{k}^t } \leq  \sum_{k\in\mathcal{K}} \sum_{t\in \mathcal{T}} \gamma_t {l}_k^t +  {\log({|\mathcal{K}|}/{\nu})}/{2}$ from Lemma $\ref{bounded_var}$.
	\end{proof}
\end{lemma}

\begin{remark}
	The parameter $\gamma_t$ serves to decrease the variance, but to increase the bias for a learning rate, $\eta_t$, resulting in a variance-bias trade-off.
\end{remark}

\begin{remark}
The parameters, $\eta_t$ and $\gamma_t$, selected irrespective of $\nu$ would entail the proposed algorithm with a high-probability bound for any confidence level $\nu$.
\end{remark}

Rearranging the variance and bias components above, the regret with respect to arm $k^*$ is bounded as follows:

\begin{proposition}\label{regret}
The cumulative regret in (3) is determined by the inequalities given in Lemma 1 and Lemma 2 for each $k^*$, and is upper-bounded by $\frac{\log |\mathcal{K}|}{\eta_{T}}+ \left(\frac{1}{2\gamma_T}+1\right)\log(|\mathcal{K}|/{\nu}) +  \sum_{t\in\mathcal{T}}   \left( \gamma_t + \frac{\eta_t}{2}\right) |\mathcal{K}|$ if $\eta_t\leq 2\gamma_t$.
\begin{proof}
	The learning regret in (3) can be decomposed into the sub-parts: $R^{T} =  R_{bias}^T +  R_{exp3}^T + R_{var}^T =  ( {L}_{k'}^T -  \tilde{L}_{}^T) + (\tilde{L}_{}^T - \hat{L}_{k^*}^T) + (\hat{L}_{k^*}^T  -  {L}_{k^*}^T) =  {L}_{k'}^T  -  {L}_{k^*}^T$ where $R_{var}^T = \hat{L}_{k^*}^T  -  {L}_{k^*}^T$ from Lemma $\ref{bounded_var}$, $R_{bias}^T =  {L}_{k'}^T -  \tilde{L}_{}^T$ from Lemma $\ref{bias}$, and $R_{exp3}^T = \tilde{L}_{}^T - \hat{L}_{k^*}^T \leq \frac{\log |\mathcal{K}|}{\eta_T} + \sum_{k\in\mathcal{K}} \sum_{t\in\mathcal{T}} \frac{\eta_t \hat{l}_k^t}{2}$ from \cite{Bubeck2012}. The upper bound of $R_{exp3}^T$ is further managed by manipulating a term of the bias, $\sum_{k\in\mathcal{K}} \sum_{t\in \mathcal{T}}   {\gamma_t  \hat{l}_{k}^t}$ with a proper value of $\eta_t$, i.e., conditioned on $\eta_t/2\leq \gamma_t$. To sum up, the aggregate regret is upper-bounded by $R^T \leq  \frac{\log |\mathcal{K}|}{\eta_T}+  \frac{\log(|\mathcal{K}|/{\nu})}{2\gamma_T} + \sum_{k\in\mathcal{K}} \sum_{t\in\mathcal{T}}  \left( \gamma_t + \frac{\eta_t}{2}\right)  \hat{l}_k^t \leq  \frac{\log |\mathcal{K}|}{\eta_T}+ \left(\frac{1}{2\gamma_T}+1\right)\log(|\mathcal{K}|/{\nu}) + \sum_{k\in\mathcal{K}}\sum_{t\in\mathcal{T}}   \left( \gamma_t + \frac{\eta_t}{2}\right) {l}_k^t \leq  \frac{\log |\mathcal{K}|}{\eta_T}+  \left(\frac{1}{2\gamma_T}+1\right)\log(|\mathcal{K}|/{\nu}) + \sum_{t\in\mathcal{T}}   \left( \gamma_t + \frac{\eta_t}{2}\right) |\mathcal{K}|$ if $\eta_t\leq 2\gamma_t$.
\end{proof}
\end{proposition}

\begin{remark}
	The learning regret can be bounded by controlling variance and bias.
\end{remark}

\begin{remark}
The step-size, $\eta_T$ and $\gamma_T$, vanishing rapidly may be sub-optimal, since it may incur a higher regret.
\end{remark}

{ 
The implicit exploration enables reliable cost estimation in assessment rule, thereby obtaining a high-probability $1-\nu$ regret bound. However, the bound is not compatible with adaptation to the dynamics in resource demand and supply, since an arm is typically selected based on the assumptions that all candidate VFNs have i) identical resource demand and ii) fair opportunity to be assessed. Next, focusing on achieving a better bound on regret with the same $1-\nu$ probability, taking into account two dynamic feeders, we show how the adaptation can be treated in an exploration bonus in the selection rule for suitability and fairness. 
}

\subsection{Dynamic resource demand}
In the following, the dynamic resource demand-based offloading decision making is studied.
An arm $k$ is selected with a probability proportional to $e^{-\mathcal{\hat{L}}_{k}^{t-1}}$ without mixing any explicit exploration term into the distribution, but with multiplying the normalized input data size $\delta^t$ with the cumulative score corresponding to the arm $k$. Note that the positive value of $\delta^t$ plays a role in making the high selection probability greater and the low selection probability lower, and determining the sensitivity of the probability a given arm is chosen over the estimated cumulative score values of alternative arms in the corresponding state. The lower the value of $\delta^t$, the less sensitive the probability of a given arm being chosen will be to the relative differences in the cumulative scores. On the other hand, high $\delta^t$ values cause choices to become sensitive to the estimated values of the various alternative arms.

One critical issue is that notwithstanding having estimated all the arms correctly, Boltzmann exploration may be able to pull sub-optimal arms prematurely or excessively. Such abrupt decisions may cause unintended consequences, i.e., cumulative importance-weighted loss estimates may become irreversible afterward \cite{cesa2017}. This is mainly due to the fact that Boltzmann exploration does not consider the uncertainty of the empirical cost estimates, i.e., large variance caused by unbiased bandit estimator with arbitrary small selection probability may result in a worse outcome. This work circumvents the issue above by guaranteeing algorithmic robustness for which bounded properties on estimation are used. High-probability bounds for adversarial bandits were provided in \cite{Auer2002} with Exp3P algorithm and in \cite{Neu2015} with Exp3IX, but limited to the surrogate regret with capability-based selection strategy.

Note that for task $t$ the probability of a dominant arm $p_{m}^t, {m\in \mathcal{K}}$ which is superior to the other arms $\mathcal{\hat{L}}_{m}^{t-1} < \mathcal{\hat{L}}_{k}^{t-1} \forall k\in \mathcal{K}$, or even has a low enough score with $\mathcal{\hat{L}}_{m}^{t-1} < \frac{1}{|\mathcal{K}|}\sum_{k\in \mathcal{K}}\mathcal{\hat{L}}_{k}^{t-1}$, increases in $\delta^t$, ${p_{m}^t} < p_{m|{\delta >0}}^t$ due to the fact that the derivative of the resulting probability $p_{m}^t$ in $\delta^t$ becomes a positive value, $\frac{\partial \Lambda_{m}}{\partial \delta^t} = \frac{ \sum_{k\in \mathcal{K}}  \left(\mathcal{\hat{L}}_{k}^{t-1} - \mathcal{\hat{L}}_{m}^{t-1} \right) \exp\left[{-(1+\delta^{t})(\mathcal{\hat{L}}_{k}^{t-1}+\mathcal{\hat{L}}_{m}^{t-1})}\right] }{(\sum_{k\in \mathcal{K}} \exp[{{-(1+\delta^{t})\mathcal{\hat{L}}_{k}^{t-1}}}] )^2}>0$. {The following proposition states that such escalation in the resulting probability of the dominant arm allows to have distinct but enhanced concatenation profiles, respectively, achieving a lower regret than the case with $\delta = 0$, equivalently $\delta^t = 0, \forall t\in\mathcal{T}$}.

\begin{proposition} \label{ref:prop2}
	When $\mathcal{\hat{L}}_{m}^{s} < \frac{1}{K-1}\sum_{k\neq m} \mathcal{\hat{L}}_{k}^{s}$, the upper bounds on variance in Lemma \ref{bounded_var} and bias in Lemma \ref{bias} with $\delta > 0$ become lower than the ones with $\delta = 0$ for all $t_o\leq s \in \mathcal{T}$ (a.s.), thereby obtaining the lower cumulative regret upper bound in Prop. \ref{regret}. 
\begin{proof}
{The proof follows from deriving improved concentration bound, and results in reduced upper bounds on variance and bias in Lemmas \ref{bounded_var} and \ref{bias} with  $\delta > 0$. In randomized selection rule, a dominant or dominated arm is chosen. Note that the probability of a dominant arm is significantly higher than the one of a dominated arm. Assume that a dominant arm's index is $m$ for task $s$, $\mathcal{\hat{L}}_{m}^{s-1} < \frac{1}{|\mathcal{K}|}\sum_{k\in \mathcal{K}}\mathcal{\hat{L}}_{k}^{s-1}$ and thus ${p_{m|{\delta = 0}}^s} < p_{m|{\delta >0}}^s$.

i) If the dominant arm is selected for task $s \in \mathcal{T}' = [1, s]\subseteq\mathcal{T}$, we could obtain the following inequality relation: $\sum_{k\in \mathcal{K}} \sum_{t\in\mathcal{T}'}     \alpha_t \hat{l}_{k}^t  > \sum_{k\in \mathcal{K}} \sum_{t\in\mathcal{T}'}     \alpha_t \hat{l}_{k|{\delta >0}}^t$ where $\alpha_t \hat{l}_{k}^t =      \frac{  \alpha_t {l}_{k}^t\cdot \mathbbm{1}{\{k = m\}}}{p_{k}^t+\gamma_t}$, $\alpha_t \hat{l}_{k|{\delta >0}}^t =     \frac{  \alpha_t {l}_{k}^t\cdot \mathbbm{1}{\{k = m\}}}{p_{k|{\delta >0}}^t+\gamma_t}$. From Lemmas 1 and 2, we have $\sum_{k\in \mathcal{K}}\sum_{t\in\mathcal{T}} \alpha_t  (\hat{l}_k^t - {l}_k^t) \leq \ln(|\mathcal{K}|/\nu)$, if $\alpha_t\leq 2\gamma_t$  for all $t$. By rearranging the inequalities above, we obtain $\sum_{k\in \mathcal{K}} \sum_{t\in\mathcal{T}'}  {\gamma_t  \hat{l}_{k}^t } = \sum_{k\in \mathcal{K}} \sum_{t\in\mathcal{T}'}   {\gamma_t  \hat{l}_{k|{\delta >0}}^t } + H_K^s \leq  \sum_{k\in \mathcal{K}} \sum_{t\in\mathcal{T}'} \gamma_t  {l}_k^t +  {\log({K}/{\nu})}/2$ where $H_K^s = \sum_{k\in \mathcal{K}} \sum_{t\in\mathcal{T}'}   {\gamma_t ( \hat{l}_{k}^t -\hat{l}_{k|{\delta >0}}^t) }$. When the value of $H_K^s$ is positive, intuitively, the upper bounds on $R_{var}^s$, $R_{bias}^s$, and $R_{exp3}^s$ in Prop. 1 become lower for any confidence level.

ii) Note that there is still non-negligible possibility that one of the dominated arm $k \in \mathcal{K}\backslash m$ is selected due to the random selection strategy, even though there exist multiple dominant arms $m$, $\hat{L}_{m}^{s-1} < \frac{1}{K-1}\sum_{k \neq m}\hat{L}_{k}^{s-1}$, and dominated arm $k \in \mathcal{K}\backslash m$ could get lower selection probability with $\delta>0$,  ${p_{k|{\delta = 0}}^s} > p_{k|{\delta >0}}^s$. Nevertheless, the estimated cost of the dominated arm $\hat{l}_k^s$ gets higher with $\delta>0$ due to the lower resulting probability and would increase its cumulative score, thereby making the section probability lower at the next round. 

To sum up, the proposed algorithm with $\delta>0$ achieves a lower cumulative regret than the one with $\delta =0$, $R_{\delta>0}^s \leq R_{\delta=0}^s$, as $s$ becomes large enough, $s\geq t_o$.
}
\end{proof}
\end{proposition}

A natural question is whether an arm with rather a good service capability compared to other arms results in a lower score, i.e., whether an arm with relatively low cumulative realized offloading cost also is allowed to form a low score parameter after performing a certain amount of tasks, which would be eventually effective in reducing variance and bias, and thus learning regret {(Prop. \ref{ref:prop2}).  The following proposition provides such rational selection behavior.}
 
\begin{proposition} \label{ref:prop3}
When $\frac{1}{K-1}\sum_{k\neq m}  \mathcal{L}_{k}^{s} \geq \mathcal{L}_{m}^{s}$, $\sum_{t}\eta_t \rightarrow \infty$, and $\gamma_t > t^{-1}$, then $ \frac{1}{K-1}\sum_{k\neq m} \hat{\mathcal{L}}_{k}^{s} \geq \hat{\mathcal{L}}_{m}^{s}$  for all $t_o\leq s \in \mathcal{T}$.
 	\begin{proof}
{The proof follows from treating the two sequences, $\frac{1}{K-1}\sum_{k\neq m} \zeta_k^t$ and  $\zeta_m^t$ where $\zeta_a^t = \hat{l}_{a}^t -  {l}_{a}^t, a\in \mathcal{K}^t$ \cite{Durrett2019}, as supermartingale difference sequences and from applying law of large numbers for supermartingale difference sequence. We show $\frac{1}{K-1}\sum_{k\neq m} \hat{\mathcal{L}}_{k}^{s} \geq \hat{\mathcal{L}}_{m}^{s}$ for task $s \in \mathcal{T}' = [1, s]\subseteq\mathcal{T}$ by contradiction.

Suppose the contrary, $\frac{1}{K-1}\sum_{k\neq m} \hat{\mathcal{L}}_{k}^{s} - \hat{\mathcal{L}}_{m}^{s} < 0$, and a measure in terms of a score of the cumulative distance between two $\mathcal{F}^t$-measurable random variables is expressed as $\frac{1}{K-1}\sum_{k\neq m} \hat{\mathcal{L}}_{k}^{s} - \hat{\mathcal{L}}_{m}^{s} = \sum_{t\in\mathcal{T}'} \eta_t \left(\frac{1}{K-1}\sum_{k\neq m} \hat{l}_{k}^t -\hat{l}_{m}^t\right)	= \sum_{t\in\mathcal{T}'} \eta_t  \left(\frac{1}{K-1}\sum_{k\neq m}  {l}_{k}^t -  {l}_{m}^t\right) +  {\sum_{t\in\mathcal{T}'} \eta_t \zeta(t)} = \sum_{t\in\mathcal{T}'} \eta_t   \left[\frac{ \frac{1}{K-1}\sum_{k\neq m}  \mathcal{L}_{k}^s -  \mathcal{L}_{m}^s }{\sum_{t\in\mathcal{T}'} \eta_t} +  \frac{\sum_{t\in\mathcal{T}'} \eta_t \zeta(t)}{\sum_{t\in\mathcal{T}'} \eta_t}\right]$ where $\zeta(t) =  \frac{\sum_{k\neq m} \zeta_k^t}{K-1} - \zeta_{m}^t$ and $\zeta_a^t = \hat{l}_{a}^t -  {l}_{a}^t, a\in \mathcal{K}^t$. 

Now, we show $\frac{\sum_{t\in\mathcal{T}'} \eta_t \zeta(t)}{\sum_{t\in\mathcal{T}'} \eta_t}\rightarrow 0$. According to the strong law of large numbers for supermartingale difference sequences \cite[Corollary 4.2]{Neely2012} \cite[Theorem 5]{Chow1965} \cite[Theorem 2]{Csorgo1968}, if the second moment of supermartingale differences is bounded, $\sum_{t=1}^{\infty} \frac{E[|\eta_t \zeta(t)|^2 |\mathcal{F}^{t-1}] }{[\sum_{t=1}^s\eta_t]^2}  <\infty$ and $\sum_{t=1}^{\infty}\eta_t \rightarrow \infty$, we get with probability $1$, $\frac{\sum_{t\in\mathcal{T}'} \eta_t \zeta(t)}{\sum_{t\in\mathcal{T}'} \eta_t}\rightarrow 0$.

Next, we focus on an upper bound to $E[|\eta_t \zeta(t)|^2 |\mathcal{F}^{t-1}]$. The parameter $\eta_t$ depends only on $t$, not $\mathcal{F}^{t-1}$, thus $E[|\eta_t\zeta(t)|^2 |\mathcal{F}^{t-1}] = \eta_t^2 \cdot E[|\zeta(t)|^2 |\mathcal{F}^{t-1}]$ where $E[|\zeta(t)|^2 |\mathcal{F}^{t-1}]$ is bounded as follows:
$E[|\zeta(t)|^2 |\mathcal{F}^{t-1}] = E[|\frac{1}{K-1}\sum_{k\neq m} \zeta_k^t - \zeta_{m}^t|^2 |\mathcal{F}^{t-1}] \leq 2E[|\frac{1}{K-1}\sum_{k\neq m} \zeta_k^t|^2|\mathcal{F}^{t-1}] + 2E[| \zeta_{m}^t|^2 |\mathcal{F}^{t-1}] \leq 4E[|\max{(\zeta_k^t)} |^2]  {\leq} 4E[| \hat{l}_{x}^t -{l}_{x}^t|^2 ]_{x= \arg_k\max{\zeta_k^t}}	\leq 8 {l}_{x}^t \left(   1/{\gamma_t^2} +1\right)$. Thus we get an upper bound as follows $E[|\eta_t\zeta(t)|^2 |\mathcal{F}^{t-1}] = \eta_t^2 E[|\zeta(t)|^2 |\mathcal{F}^{t-1}]\leq 8 l_x^t  \eta_t^2 \left(   \frac{1}{\gamma_t^2}  +  1 \right)$.  

By using the bound, $\lim_{s\rightarrow\infty} \sum_{t=1}^{s} \frac{E[|\eta_t\zeta(t)|^2 |\mathcal{F}^{t-1}] }{[\sum_{\tau=1}^t\eta_{\tau}]^2} \leq  \lim_{s\rightarrow\infty} \sum_{t = 1}^{s} \frac{8 l_x^t \eta_t^2\left(   {1}/{\gamma_t^2}  +  1 \right)}{[t\eta_t]^2} \leq  8 \cdot  \mathcal{O} \left(\sum_{t= 1}^s \frac{1/\gamma_t^2 + 1}{t^2}\right) < \infty$, when $t^2 \gamma_t^2 > 1$.  Thus, $\frac{1}{K-1}\sum_{k\neq m} \hat{\mathcal{L}}_{k}^{s}  - \hat{\mathcal{L}}_{m}^{s}$ is in contradiction with the nonnegativity of $\frac{1}{K-1}\sum_{k\neq m}  {\mathcal{L}}_{k}^{s} -  {\mathcal{L}}_{m}^{s}$ for all $s \geq t_o$, $\frac{1}{K-1}\sum_{k\neq m}  {L}_{k}^{s} \geq  {\mathcal{L}}_{m}^{s}$.} 
\end{proof}
\end{proposition}

\begin{corollary}
The condition $\frac{1}{K-1}\sum_{k\neq m}  \mathcal{L}_{k}^{s} \geq \mathcal{L}_{m}^{s}$  is sufficiently satisfied with $ \mathcal{L}_{k}^{s} \geq \mathcal{L}_{m}^{s} \forall k$, $\gamma_t > t^{-1}$, and $\sum_{t}\eta_t \rightarrow \infty$. 
\end{corollary}

\begin{remark}
	The step-size $\gamma_t$ vanishing faster than or equal to $ t^{-1}$, $\gamma_t \leq t^{-1}$ could be sub-optimal.
\end{remark}

\subsection{Dynamic resource supply}
In the following, we show that allowing a certain arm to get some fixed extra information at the beginning of the learning interval \cite{Duffie1986} would result in better performance than the conventional approach, i.e., partial or full reset in Section III-B.
Such fine-tuned scores enable to reduce the exploration space, thereby rapidly calibrating perception and adapting to environmental changes. One may share the explored information among the arms which follow the same cost distribution for reducing the exploration space, but only valid in stochastic MAB framework \cite{Zhang2020}. {Considering the adversary's non-stochastic force, when the corresponding arm reappears after a finite but not so many rounds, a task requester may reuse its own previous score result, or may use the minimum of the other arms' scores in the immediately previous round if no exploration progress has been made for many rounds, $\beta_{\bar{k}} = \max(\min(\mathcal{\hat{L}}_{m}^{\tau-1}),\mathcal{\hat{L}}_{\bar{k}}^{\tau-1}), m\in \underline{\mathcal{K}}^{\tau-1}$.}  If an arm disappears in round $\tau$ due to its inherent mobility, a task requester could alleviate the unnecessary pull by ruling out the vanishing arm in its selection process, $|\mathcal{K}^{\tau}|< |\mathcal{K}^{\tau-1}|$.

We compare the proposed approach for volatile resource supply case where the arm $\bar{k}$ which recently joined the exploration process for previous task $\tau_{i,p}$ appears again for the current task ${\tau_i}=\mathcal{T}_i[1]\in \mathcal{T}_i$, with another two cases: i) zero value of $\beta_{\bar{k}}$ for the partial reset case, i.e., only $\hat{L}_{\underline{k}}^{\tau_i-1}>0$, and ii) zero values of $\beta_{\bar{k}}$ and $\hat{L}_{\underline{k}}^{\tau_i-1}$ for the full reset case. {The parameter $\beta_{\bar{k}}$ is not iteration-varying but interval-varying, i.e., updated whenever the candidate fog node set is changed for task $\tau_i, \forall i$. 
Note that while the modified score for dynamic demand affects all arms equally, the one for dynamic supply only tunes the scores of joining arms, which results in unfair exploration and different filtration processes.  Denote $\beta_{\bar{k}} >0 \forall i$ by $\beta >0$.} In the following, we show that $\beta_{} >0$ allows to have enhanced learning performance with a better composite of cumulative scores over arms.

\begin{proposition} \label{ref:prop4}
	When $\sum_k\hat{\mathcal{L}}_{k}^{s}>\sum_k\hat{\mathcal{L}}_{k|{\beta>0}}^{s}$ where $k\in \mathcal{K}^{\tau_i}, s \in \mathcal{T}_{i}~\forall i$, the lower variance and bias are obtained.
	\begin{proof}	
		{The proof follows from deriving improved concentration. By considering task $t \in \mathcal{T}_{i} = [\tau_i, s]$ and arm $k\in\mathcal{K}^{\tau_i}$ and modifying the term $\tilde{\lambda}_t = \sum_{k} \eta_t    \tilde{l}_k^t$ in Lemma \ref{bounded_var} with $\tilde{\lambda}_{t|{\beta  >0}} = \sum_{k} \eta_t \tilde{l}_{k|{\beta  >0}}^t $, if $\eta_t\leq 2\gamma_t$, one gets $E[Z_{\beta  >0}^{t} | \mathcal{F}^{t-1}]\leq Z_{\beta  >0}^{t-1}$ where $Z_{\beta  >0}^{t} = \exp(\sum_{t} (\tilde{\lambda}_{t|{\beta  >0}} - {\lambda}_t))$ remains also super-martingale relative to $\mathcal{F}^{t-1}$.} With the respective complement probability $1- v$ and bandit feedback, one gets $\sum_{t}\sum_{ {k}} \eta_t  (\hat{l}_{ {k}}^t - {l}_{ {k}}^t)  = \sum_{t}   \sum_{{k}} \eta_t (\hat{l}_{ {k}|{\beta  >0}}^t -  {l}_{ {k}}^t) + H_{{K}}^s  \leq \ln(|\mathcal{K}_i|/\nu)/2$ where $H_K^s =  \sum_{t}    \sum_{ {k}}\eta_t (\hat{l}_{ {k}}^t - \hat{l}_{ {k}|{\beta   >0}}^t) $. According to Lemmas 1 and 2, when $\sum_{k}\hat{\mathcal{L}}_{k}^{s}>\sum_{k}\hat{\mathcal{L}}_{k|{\beta  >0}}^s$, one gets the lower variance and bias.
	\end{proof}
\end{proposition}

The result above enables a balance of the exploration process needed to identify the reasonable alternative in a fair manner, which results in lower learning regret with using Prop.~$\ref{regret}$, conditioned on a better formed of cumulative scores. Now, the natural question is whether we have $\sum_{k\in\mathcal{K}_i} \hat{\mathcal{L}}_{k}^{s} > \sum_{k\in\mathcal{K}_i}\hat{\mathcal{L}}_{k|{\beta_i  \uparrow}}^s, s \in \mathcal{T}_{i}, \forall i$. The following proposition states that the cumulative score of the proposed approach is better than the ones of the reset cases.

\begin{proposition} \label{ref:prop5}
When $\gamma_t / \eta_t = \phi > 0~\forall t$, $\sum_k\hat{\mathcal{L}}_{k}^{s}>\sum_k\hat{\mathcal{L}}_{k|{\beta >0}}^s, k\in \mathcal{K}^{\tau_i}, s \in \mathcal{T}_{i}, s \geq t_o, \forall i$.
	\begin{proof}		
		{The proof follows from the use of the filtration associated with stochastic process, showing that positive values of the modified score and implicit exploration parameters would ensure improved concentration.
	 
		Suppose the contrary, $\sum_k\hat{\mathcal{L}}_{k}^{s} \leq \sum_k\hat{\mathcal{L}}_{k|{\beta  >0}}^s, k\in \mathcal{K}^{\tau_i}$ where $\sum_k\hat{\mathcal{L}}_{k}^{s}$ and $\sum_k\hat{\mathcal{L}}_{k|{\beta >0}}^{s}$ are from two different filtration sets $\mathcal{F}^{s-1}$ and $\mathcal{F}_{{\beta >0}}^{s-1}$ of the measures used in estimation, respectively. A filtration represents iteration-varying available knowledge, an increasing sequence of sigma algebras, i.e., $\mathcal{F}^1 \subseteq \cdots \subseteq \mathcal{F}^{s-1}$ and $\mathcal{F}_{{\beta >0}}^1 \subseteq \cdots \subseteq \mathcal{F}_{{\beta >0}}^{s-1}$ where $\mathcal{F}^{s-1}$ and $\mathcal{F}_{{\beta >0}}^{s-1}$ are information available for task $s$.  A larger information would allow to provide a more accurate estimate. The amount of information is different in the two filtration sets, because joining arms (partial) or all arms (full) initiate exploration process with lack of information. Using a modified score $\beta>0$  plays a role in getting extra information for task $\tau_i$ influencing the subsequent estimates for tasks $s> \tau_i$. 
		 
		From the two filtration sets $\mathcal{F}$ and $\mathcal{F}_{{\beta \uparrow}}$, we compare $\sum_k\hat{\mathcal{L}}_{k|{\beta>0}}^s$ with $\sum_k\hat{\mathcal{L}}_{k}^{s}$ under the partial and full reset cases.   
		
		i) for the partial reset case, $\mathcal{\hat{L}}_{\bar{k}}^{\tau_i-1} = 0$ and $\beta = 0$, one may consider a certain task offloading round, $\tau_{i}'$ as a self-adjoint operator adapted to the filtration, representing the number of additional exploration rounds for which the newly or reappeared arms need to experience to become comparable, but in fact could save via $\beta>0$ where $\beta = \sum_{t} \eta_t \hat{l}_{\bar{k}}^t  \leq \sum_{t }  \frac{{l}_{\bar{k}}^t}{ \gamma_t / \eta_t}, t \in [\tau_i, {\tau_i + \tau_{i}'-1}]$. When $ \gamma_t / \eta_t$ is fixed over task rounds, $\gamma_t / \eta_t = \phi>0 \forall t$, the least number of the saved explorations is positive, $\tau_{i}' \geq  \frac{\phi \beta}{\max({l}_{\bar{k}}^t)}>0$, with $\mathcal{F}_{}^{\tau_{i}'} \subseteq \mathcal{F}_{{\beta>0}}^{\tau_{i}'}$.

		ii) for the full reset case, $\hat{L}_{{k}}^{\tau_i-1} = 0$ and $\beta= 0$, taking into account the property of  importance-weighted sampling, one considers the fine-tuned scores substracting the minimum of the existing arms' scores, $\hat{\mathcal{L}}_{k|{\beta>0}}^{\tau_i-1} - \min(\hat{\mathcal{L}}_{\underline{k}}^{\tau_i-1})$, but the score deviation among the existing arms still remains and addresses the distinction of their offloading capabilities saving positive explorations as in the partial reset, with $\mathcal{F}^{\tau_{i}-1} \subseteq \mathcal{F}_{\beta>0}^{\tau_{i}-1}$. 
		
		To sum up, for the both cases, the fine-tuned scores allow to have $\sum_k\hat{\mathcal{L}}_{k}^{s} > \sum_k\hat{\mathcal{L}}_{k|{\beta>0}}^s, k\in \mathcal{K}^{\tau}$ which contradicts.  }  
	\end{proof}
\end{proposition}
{From the result above, the proposed approach $\beta>0$ permits to save up to at least a positive value of exploration rounds with a positive implicit exploration parameter. }

\subsection{Sub-linear regret}
A learning algorithm is said to achieve the {no-regret} condition if the cumulative regret has a sub-linear growth rate with the number of task, $T$, in other words, the per-round regret is vanishing \cite{cesa2006}, i.e., negligible as $T$ grows, $ {R^T}/{T} \rightarrow 0$. Note that the parameters, $\eta_t$ and $\gamma_t$, of a potential fog node existing for the multi-intervals, $i\in \mathcal{I}$ are decreasing with respect to $t\in \mathcal{T}$, i.e., iteration-varying, while the candidate set might be different for different interval, i.e., interval-varying.

\begin{proposition} \label{ref:bound}
	When $\eta_T > \frac{1}{T}$ and $\gamma_T > \frac{1}{2T}$, the cumulative regret has sub-linearity.	
	\begin{proof}
	The upper bound on the per-round regret from Prop. \ref{regret} for the multi-intervals $i\in\mathcal{I}$ is, $ \sum_{i\in \mathcal{I}} \sum_{t\in\mathcal{T}_i}   \left( \gamma_t + \frac{\eta_t}{2}\right) \frac{|\mathcal{K}_i|}{{T}_i} + \sum_{i\in\mathcal{I}} \left( \frac{1}{{T}_i}\frac{\log |\mathcal{K}_i|}{\eta_{{T}_i}}+ \frac{1}{{T}_i}\left(\frac{1+ 2\gamma_{{T}_i}}{2\gamma_{{T}_i}}\right)\log(|\mathcal{K}_i|/{\nu})\right)$ where the first term vanishes w.r.t ${T}_i = |\mathcal{T}_i|$ and the second term also vanishes if $\eta_{{T}_i}\leq 2\gamma_{{T}_i}$, in particular $\eta_{{T}_i} > \frac{1}{{{T}_i}}$ and $\gamma_{{T}_i} > \frac{1}{2{|\mathcal{T}_i|}}, \forall i \in \mathcal{I}$. Consequently, the overall cumulative regret has sub-linearity, $\lim\sup_{T\rightarrow \infty}  {R^T}/{T}  \leq 0$ if the one for individual interval has sub-linearity.	
	\end{proof}
\end{proposition}

\begin{proposition} \label{ref:bound}
	The per-round regret incurred by the proposed algorithm vanishes faster than those on the undynamic case.	
	\begin{proof}		
		Note that when the individual sub-parts in Prop. \ref{regret} have sub-linearity respectively, then it is sufficiently satisfied that the considered $R^T$ has sub-linearity, and furthermore stronger sub-linearity is allowed by considering the dynamic resource demand and supply in the proposed algorithm as discussed in Sec. IV-B and IV-C which make the upper bounds of the individual sub-parts in Prop. 1 lower, meaning (lower and) faster vanishing regret.
\end{proof}
\end{proposition}

\begin{remark}
	The conditions on the algorithm's step-size, $\eta_T$ and $\gamma_T$, allow to avoid a sub-optimal exploration-exploitation balance.
\end{remark}

\begin{corollary}
	If the learning rate is the candidate set and task round dependent in a form of $\eta_t = \sqrt{\frac{\log |\mathcal{K}_i|}{|\mathcal{K}_i| \cdot t}}$ \cite{Bubeck2012}, the algorithm’s sub-linearity properties are in effect after at least $\lceil \frac{|\mathcal{K}_i|}{\log |\mathcal{K}_i|} \rceil$ rounds.
	\begin{proof}		
	With a decreasing but candidate set dependent schedule, $\eta_t = \sqrt{\frac{\log |\mathcal{K}_i|}{|\mathcal{K}_i| \cdot t}} > t^{-1}$, the number of the candidate fog nodes needs to be larger than $|\mathcal{K}_i| >  e^{-\mathcal{W}_{-1}(t)}$ where $\mathcal{W}(\cdot)$ is Lambert function for $t$. Likewise, once the available set $|\mathcal{K}_i|$ is updated, the iterative procedures should be larger than the minimum task rounds $\tau_{i,o}$ to ensure the sub-linearity of the proposed algorithm, $\tau_i > \tau_{i,o}$ where $\tau_{i,o} = \frac{|\mathcal{K}_i|}{\log |\mathcal{K}_i|}$. Note that $\tau_{i,o}$ is increasing in $|\mathcal{K}_i| > e$ due to $\frac{\tau_{i,o}}{\partial |\mathcal{K}_i|} = \frac{\log (|\mathcal{K}_i|) -1}{\log^2 (|\mathcal{K}_i|)}$ and the lower bound on $\tau_{i,o}$ is $e$ for $|\mathcal{K}_i|= e$, i.e., $\tau_{i,o} \approx 2.9$ for $|\mathcal{K}_i| = 2$ or $\tau_{i,o} \approx 2.7$ for $|\mathcal{K}_i| = 3$. 
    \end{proof}
\end{corollary}

\section{Numerical illustration}
 
This section conducts numerical studies to assess the average per-bit cost (bit-cost) and regret of the proposed algorithm.

\begin{figure}[t]
	\centering
	\includegraphics[width=0.475\textwidth]{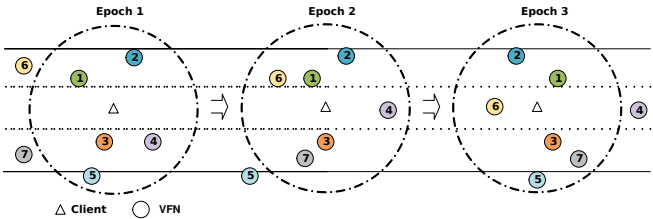}
	\caption{A VFC simulation scenario with 1 client and 7 volatile VFNs.}\vspace{-1em}
	\label{scenario}
\end{figure}

\subsection{Performance evaluation}  
\subsubsection{{Evaluation setting}} 
Consider one vehicle (client) of interest, requesting the computational resource from candidate edge computational resource providing vehicles (VFC nodes). The distance between the client and each candidate VFC node is assumed to follow a uniform distribution, $d\sim\mathcal{U}[0,d_r]$ where $d_{r}$ is the communication range equal to 400 m. The transmission power of the client is $24$ dBm, the large-scale fading gain follows the 3GPP pathloss model~\cite{3GPP2011}, $A_o = 128.1 + 37.6\log_{10}(d)$, the small-scale fading gain follows Rayleigh distribution with unit variance, channel bandwidth is $W = 10$ MHz, and noise power is $N_o = -174$ dBm/Hz. Note that the interference effects on the co-channel and adjacent channel are assumed to be ignored according to the orthogonal channel allocation \cite{Kenney2011} and experimental result \cite{Rai2007}. Also, one assumes that the service discovery solution which finds neighboring VFCs within the client's communication range allows to select fog nodes in the same moving direction as candidates \cite{Zhang2020}. Thus, the small relative speed makes the Doppler shift insignificant and fading gains remain unchanged during the uplink transmission for each task offloading request.

Consider 7 volatile VFNs (see Fig. \ref{scenario}) with maximum CPU frequency, $F_k \in \{6, 4, 5, 4, 1.5, 2, 4\}$ GHz that appear or disappear as candidate fog nodes of one task generating vehicle (client) for a finite number of tasks in 3 epochs, within each epoch consisting of 1000 tasks and keeping the same fog node set. In the first epoch, there are 5 candidate VFNs, $\mathcal{K}_t= \{1, 2, 3, 4, 5\}, \forall t\in [1, 1000]$. At the beginning of the second epoch, a less powerful VFN 5 disappears and VFNs 6 and 7 with higher computing capability appear, $\mathcal{K}_t= \{1, 2, 3, 4, 6, 7\}, \forall t\in [1001, 2000]$. At the beginning of the third epoch, VFNs 4 disappear, while VFN 5 re-appears, $\mathcal{K}_t= \{1, 2, 3, 5, 6, 7\}, \forall t\in [2001, 3000]$. For each VFN, the allocated CPU frequency to the task client is a fraction of the maximum CPU frequency which is distributed from 20\% to 50\%, but arbitrarily constrained. To address such a non-stochastic environment,
adversarial perturbation is considered in a similar manner as in \cite{Zimmert2019}, where the realized cost function is affected by the oblivious attack, specifically an arbitrary fraction of allowable CPU frequency range. The total tasks are splitted into phases with different lengths, each of which is with different means for different arms. The computation intensity is set to $w = 1000$ Cycles/bit. To meet the client's diverse demand, the request service type can be changed with different task size arbitrarily. Varying service types could be considered at regular intervals. For simplicity, a periodic interval for changing service types is aligned with an epoch. The task size, $\delta$ Mbits, is either fixed or randomly distributed according to either uniform or truncated normal distribution on a predefined interval $\delta \in [0.2, 1]$.  

\begin{figure}[t]
	\centering
	\begin{tabular}{c}
		\subfigure[\label{numerical:all:regret}]
		{\includegraphics[width=0.40\textwidth]{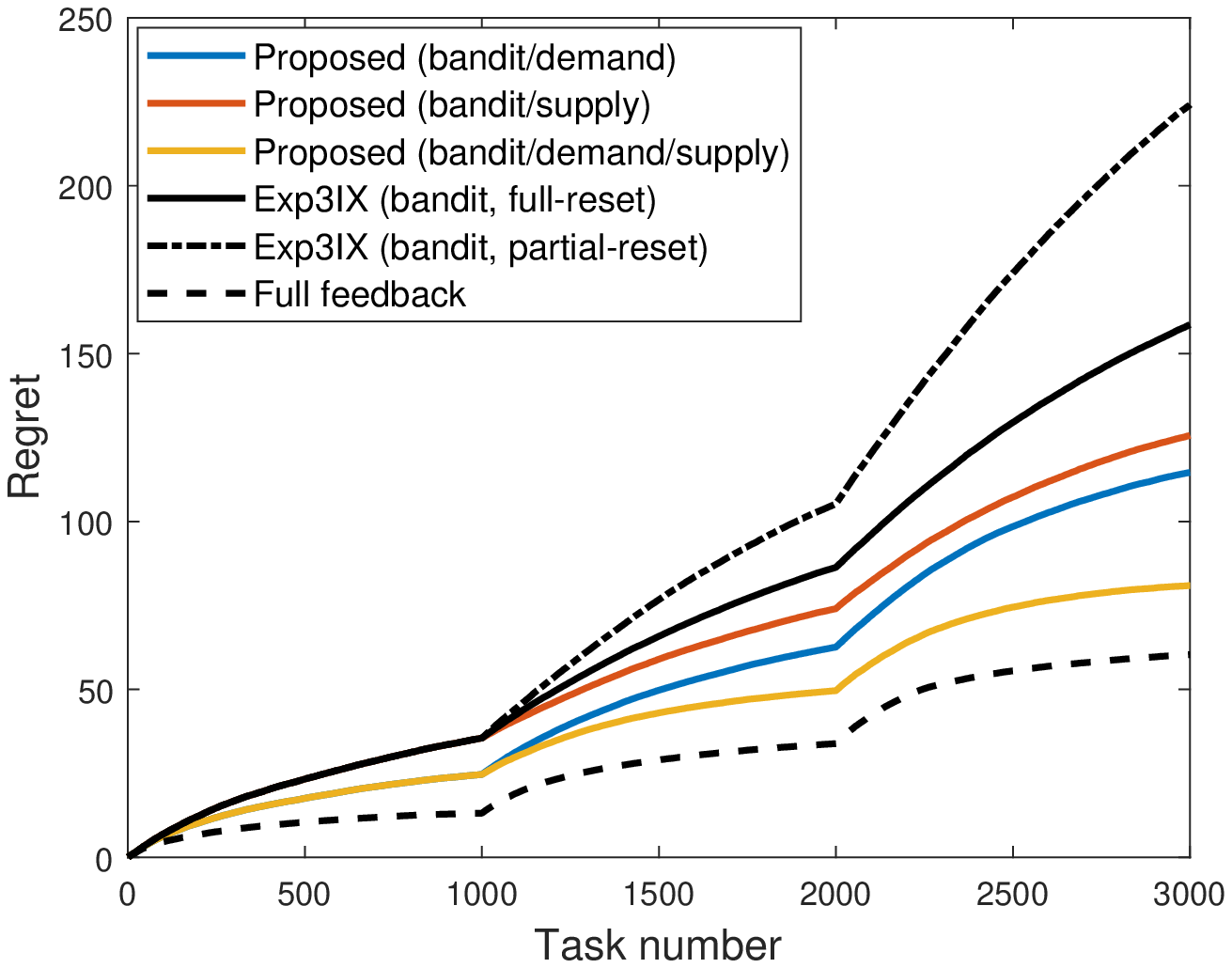}} \\
		\subfigure[\label{numerical:all:latency}]
		{\includegraphics[width=0.40\textwidth]{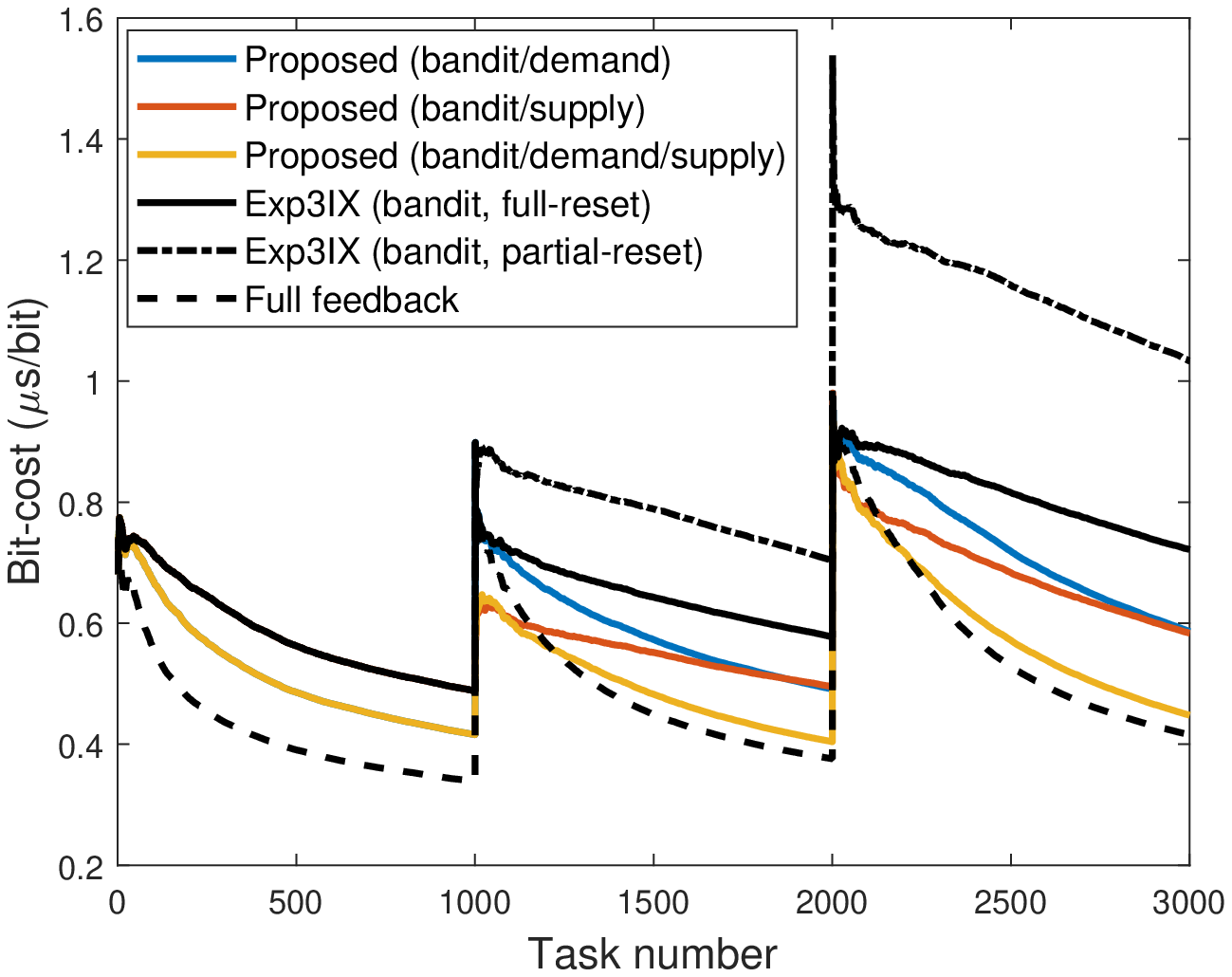}} 
	\end{tabular}\vspace{-.1em}
	\caption{Impact of dynamic resource demand ($\delta>0$) and supply ($\beta>0$) dependent fog node selection policy on the average per-bit performance (in multi-epochs): (a) regret and (b) bit-cost.}
	\label{numerical:all}
\end{figure}

%

\subsubsection{{Evaluation result}}
The proposed algorithm is compared with its counterparts, implicit exploration-based algorithms with bandit feedback and full-feedback. The performance results of learning algorithms in terms of the learning regret and the average per-bit cost when $\xi= 1$ in equation~(\ref{general_utility}), per-bit latency, are depicted in Fig. \ref{numerical:all}, showing that the proposed algorithm outperforms other implicit exploration-based algorithms where an arm is selected based on the scores i) fully reset with zero values of $\beta_i$ and $\hat{L}_{\underline{k}}^{\mathcal{T}_i[1]}$ (full-reset), and ii) partially reset with zero value of $\beta_i$ (partial-reset). Two kinds of adaptivity including dynamic resource demand and supply are considered, and notably the joint consideration of these two aspects could achieve a better exploration-exploitation trade-off {since they allow to adapt to the dynamic task offloading environment without exploring the sub-optimal actions,} thereby reducing the  regret by 65\% and 40\% from that of two conventional Exp3IX based variants, respectively, and being much closer the full information  setting \cite{Bubeck2012} where the complete cost vector is revealed after every round (full feedback) in Fig. \ref{numerical:all:regret}.

\begin{figure}[t]
	\centering
	\includegraphics[width=0.40\textwidth]{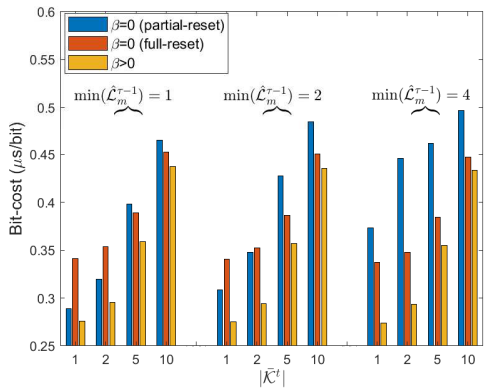}
	\caption{Average per-bit cost performance w.r.t the number of VFNs.}\vspace{-.1em}
	\label{numerical:dynamic1}
\end{figure}

Also, the proposed algorithm offers the sub-linearity of the regular regret performance, i.e., the regret grows sub-linearly with respect to the number of tasks, intuitively indicating that the task generating client's learning algorithm allows to asymptotically converge to the VFC node with optimal performance. Note that in the first epoch, the algorithm with dynamic supply and demand is equivalent to the one with only dynamic demand, and the one with only dynamic supply is equivalent to the Exp3IX. Besides, in the optimal genie-aided policy, the client always connects to the VFC node with minimum cost. As shown in Fig.~\ref{numerical:all:latency}, during each epoch, the average per-bit cost of the proposed algorithm converges faster than other ones, except for the full-feedback case.

%
%
%


\subsubsection*{\textbf{Impact of $\beta$}}
Specifically, implicit exploration-based algorithms taking into account dynamic resource supply could achieve lower learning regret compared to vanilla Exp3IX algorithms with initializing the learning history of all candidates, $\hat{L}_{k} = 0, \forall k$ or a new one $\hat{L}_{n} = 0$ whenever a candidate fog node set is updated. When a client discovers a newly appeared VFC candidate, its weight is set to the lowest one of the other candidates. For example, VFC 6 and 7 nodes appearing at $\tau = 1001$ are initialized with the score lastly updated by a rather more capable fog node, $\min(\hat{L}_k^{\tau-1})$. If a VFC leaves a candidate set temporarily but returns into the set in a finite number of tasks, the MIX-AALTO utilizes the information on the lastly updated score the rejoining VFC node had before or the other fog nodes have. For example, VFC 5 node reappearing at $\tau = 2001$ may launch its score value with the one at $\tau_p = 1000$ or $\min(\hat{L}_k^{\tau-1})$ depending on the circumstance. Such dynamic resource supply-based policy ensures that a newly discovered or re-discovered VFC is likely to be explored such that the proposed algorithm allows to avoid unfair selection opportunities and thus adapt quickly to the change in a volatile environment. This indicates that dynamic resource availability-based policy draws better adaptivity to the dynamic and adversarial environments, and thus reduces loss of performance through learning.




{Fig.\ref{numerical:dynamic1} shows the impact of the number of VFNs, $|\mathcal{\bar{K}}^t|$, appeared in the candidate set,  $k\in {\mathcal{K}}^t =  \underline{\mathcal{K}}^t\cup \bar{\mathcal{K}}^t$ for task, $t \in [1001, 2000]$ where $\mathcal{\underline{K}}^t = \{1, 2, 3, 4\}$ are the existing VFNs from the first epoch $t \in [1, 1000]$, and $\mathcal{\bar{K}}^t$ is the appearing VFNs whose distances to the client and CPU frequencies randomly selected from $\mathcal{U}$ and $F_k$. As the density of the candidate VFNs becomes higher, more exploration would be performed, requiring more rounds to make the unit offloading cost converged and resulting in a higher regret. This observation would give an implication to the design of the discovery process protocol. For instance, limiting the maximum allowable number of the candidate VFNs would be beneficial when the service requirement is strict or the network topology has a high degree of volatility. One may adjust the maximum number of candidate VFNs properly~\cite{Feng2017}. The proposed algorithms outperform the other two exploration reset cases; modifying cumulative scores only for the appearing VFNs (partial-reset) and for all VFNs (full-reset). Compared to the partial-reset case, the better the performance gain of the proposed dynamic supply based algorithm  ($\beta>0$) is achieved, the larger the minimum gap to the existing arms' scores is obtained after the task $\tau-1$, $\min(\mathcal{\hat{L}}_{m}^{\tau-1}), m\in \underline{\mathcal{K}}^{\tau}$ where $\tau = 1001$. This is because the proposed dynamic supply approach would allow to reduce the exploration rounds the appearing arms may require to experience. 
Compared to the full-reset case, on the other hand, the effect of the minimum gap to the existing VFNs' cumulative scores on the performance gain is minimal, since the score differences among the existing VFNs are only effective in distinct filtration set. Such residual difference would influence the estimation performance. On the other hand, a high density of the appearing VFNs may alleviate the effect of such score deviations among the existing VFNs, since an importance weighted mechanism assigns a probability proportional to the number of the candidate VFNs as well as the cumulative scores. 
}

\begin{figure}[t]
	\centering
	\begin{tabular}{c c } \hspace{-20pt} 
		\subfigure[\label{numerical:demand:benchmark}]
		{\includegraphics[width=0.27\textwidth]{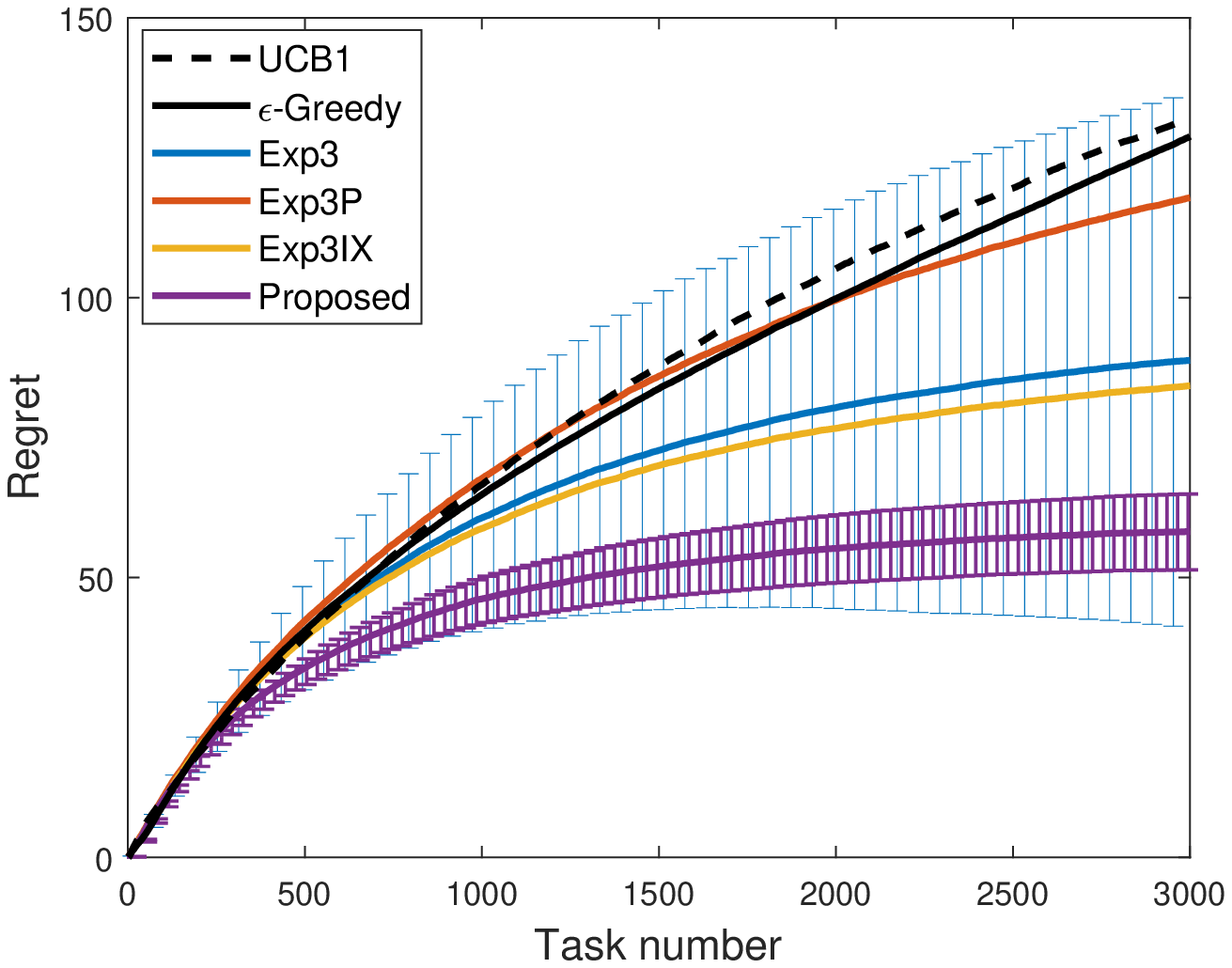}} &
		\hspace{-20pt}
		\subfigure[\label{numerical:demand:regret}]
		{\includegraphics[width=0.27\textwidth]{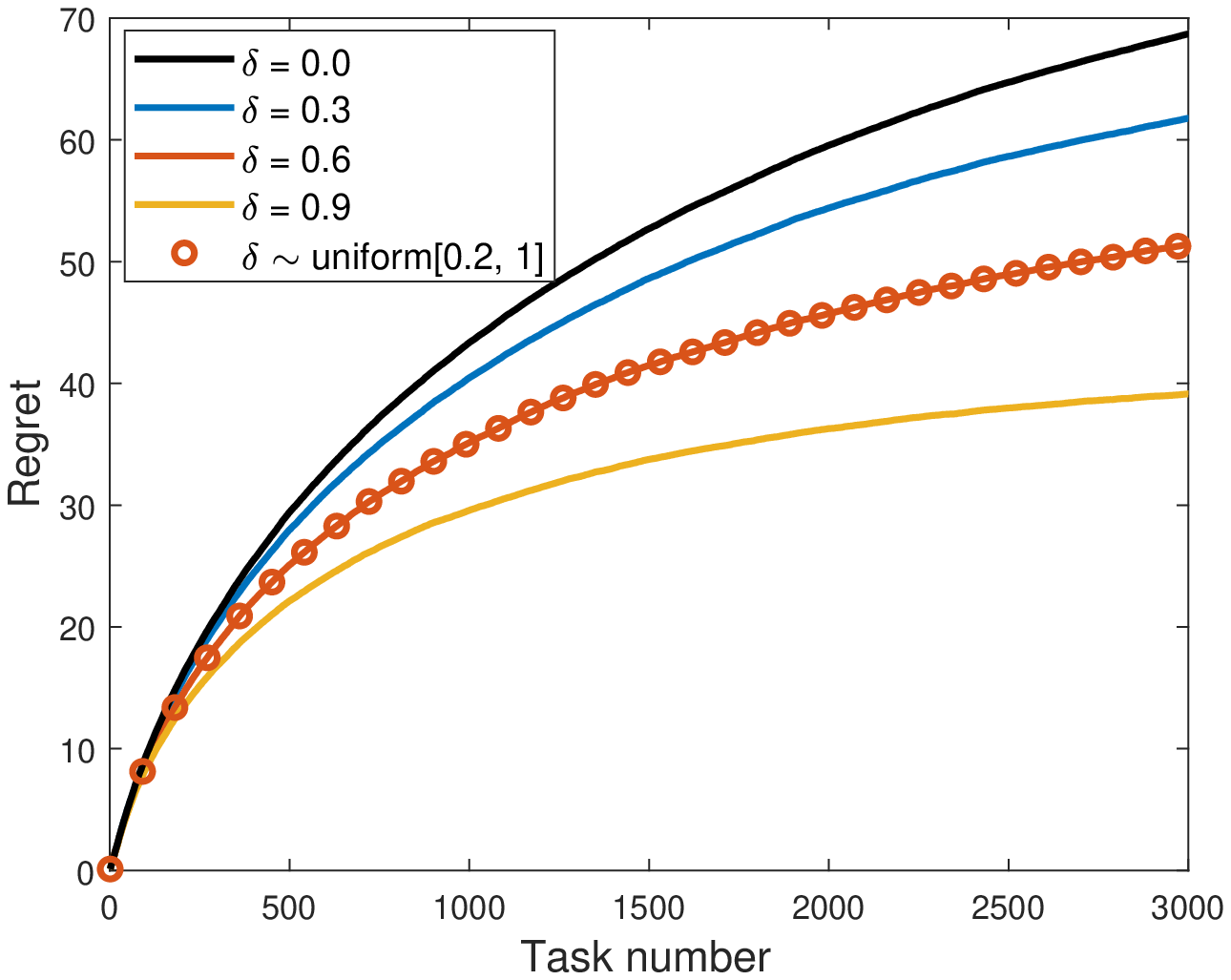}} \vspace{-0.5em}\\ \hspace{-20pt} 
		\subfigure[\label{numerical:demand:variance}]
		{\includegraphics[width=0.27\textwidth]{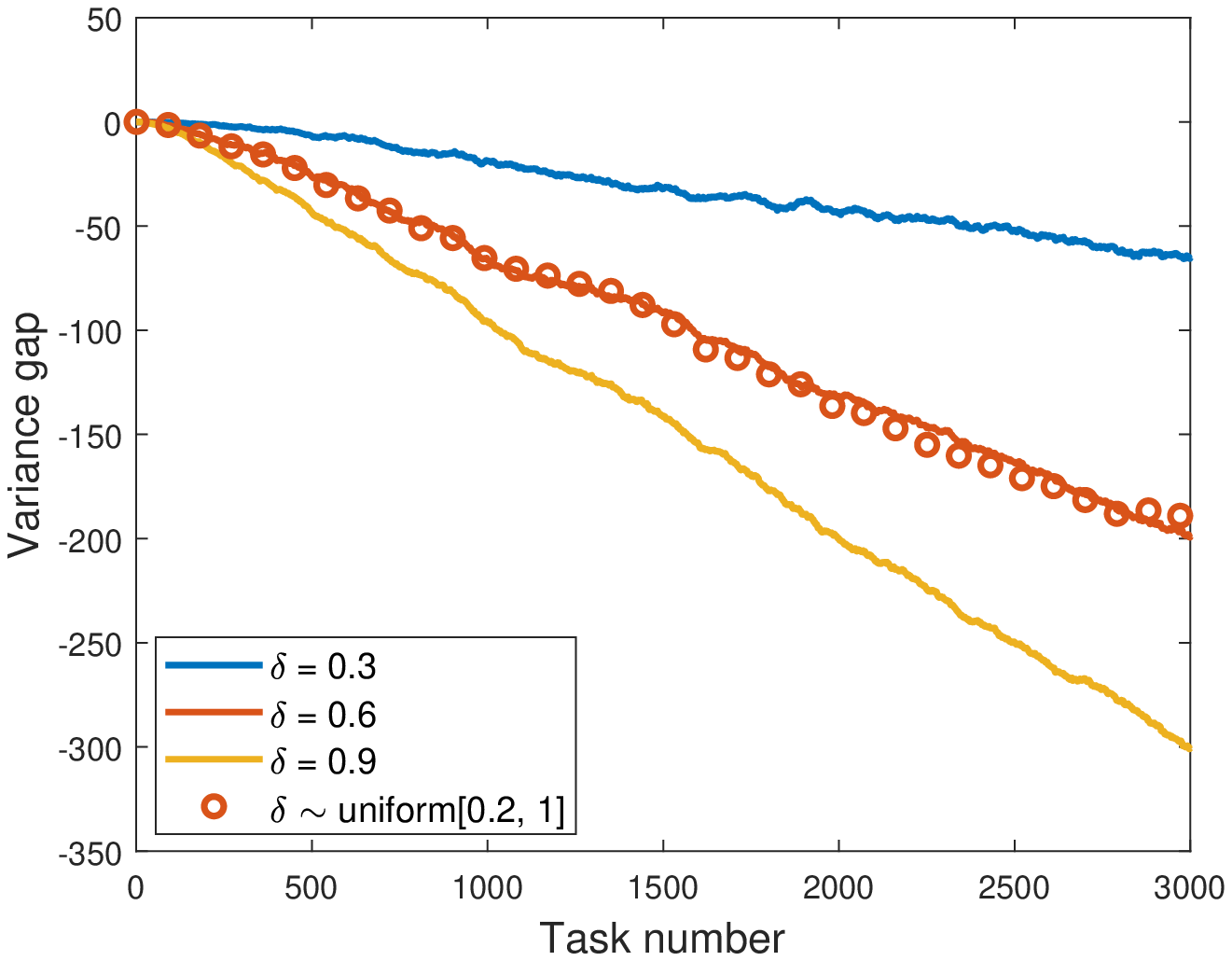}} &
		\hspace{-20pt}
		\subfigure[\label{numerical:demand:bias}]
		{\includegraphics[width=0.27\textwidth]{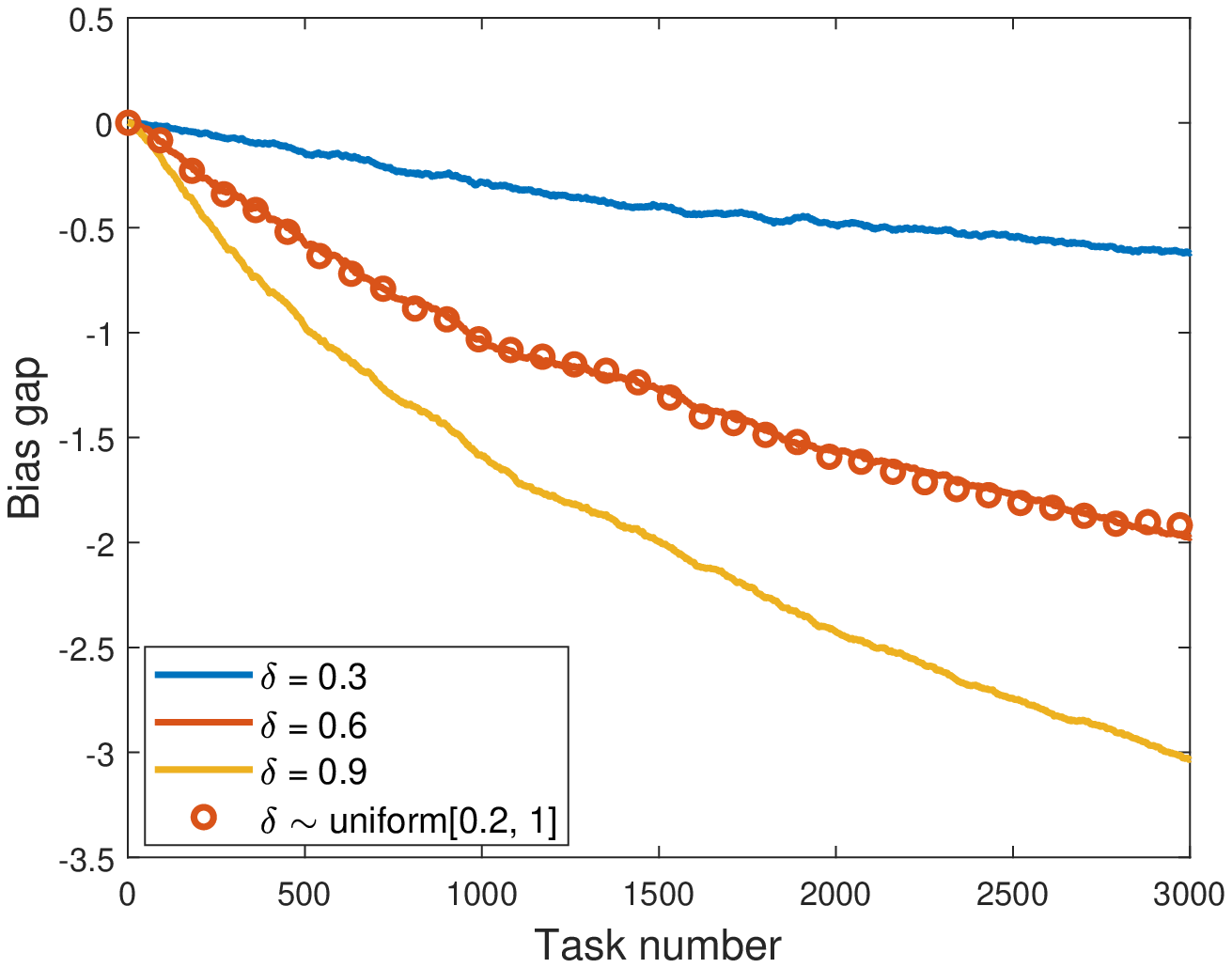}} \\
	\end{tabular}\vspace{-.1em}
	\caption{	
		(a) Regret performance comparison among UCB1\cite{Auer2002-2}, $\epsilon$-Greedy, Exp3, Exp3P and Exp3IX, when base parameters are set as in \cite[Theorems 3.1 and 3.3]{Bubeck2012} and (b) impact of dynamic resource demand dependent selection policy on the learning regret performance. And differences between these two kinds of cases, when $\delta>0$ vs $\delta=0$, in terms of (c) variance and (d) bias. 
	}\vspace{-0.35em}
	\label{numerical:demand}
\end{figure}

%
%


\subsubsection*{\textbf{Impact of $\delta$}}
For the dynamic resource demand, the proposed algorithm considers two major exploration perturbations: one is implicit exploration for guaranteeing low-variance in assessment rule and another is Boltzman exploration for drawing suitability-based selection. Fig. \ref{numerical:demand:benchmark} demonstrates the robustness of the proposed algorithm as compared to Exp3 and its superior performance as compared to other algorithms choosing arms based on current knowledge with a probability $1-\epsilon$ such as $\epsilon$-Greedy when $\epsilon = 0.1$, upper conﬁdence bound such as UCB1, and guaranteeing high probability bounds such as Exp3P and Exp3IX. This clearly shows that a fine-grained implicit exploration approach could achieve higher and more robust performance, lower empirical mean and standard deviation of the regret than others. 

%
%
%
%

Fig.~\ref{numerical:demand:regret} shows the effect of suitability-based selection policy on the learning regret. In general, when a positive value of the normalized input data size, $\delta>0$, is considered in a selection rule, a client's learning performance can be improved. This means that considering a score associated with both normalized per-task cost and per-bit cost, would make a more suitable candidate and thus ensure a better trade-off between exploitation and exploration. On the other hand, when $\delta = 0$, there is no exploration for the suitability, but only for the capability of candidate VFC nodes. Since such a capability-based learning approach may fail to address appropriately the upcoming variations of computational demand, the learning regret for the per-bit cost is explicitly worse than those of $\delta>0$. This can be captured in the learning regret with per-bit latency cost function, i.e., when $\xi = 1$ in equation~(\ref{general_utility}).
 
The effect of different input data size\footnote{While task workload is determined by the task size and computation intensity, the bit cost is only dependent on the computation intensity.} on the learning regret is also evaluated with three fixed and one uniformly distributed sizes ranged between 0.2 and 1 Mbits. The proposed algorithm brings better performance gain by making exploitation more for a large $\delta$ and less for a small $\delta$. The gain becomes larger as the input data size increases. The per-bit learning regret is reduced by around 15\%, 30\%, and 45\% from that of 0.3 Mbits, 0.6 Mbits, and 0.9 Mbits, respectively, only considering the capability-based selection policy. This observation reveals the vital role of the proposed algorithm in coping with dynamic resource demand, which is enabled by variance and bias reduction techniques in Section IV-B and IV-C. Corresponding diminishing effects of variance and bias can be captured in Fig.~\ref{numerical:demand:variance} and Fig.~\ref{numerical:demand:bias}. Apparently, when a user with a task of large size selects a VFC with weak service capability, it yields poorer learning performance than the case with a task of small size. On the other hand, selecting a low-capable VFC for a small input data size does not yield enormous delay. With varying input uniformly distributed over the same range with the fixed one, one may yield a similar result with the fixed one using the mean value of 0.6.

\begin{figure}[t]
	\centering
	\begin{tabular}{c c } \hspace{-15pt} 
		\subfigure[\label{numerical:ee_latency}]
		{\includegraphics[width=0.260\textwidth]{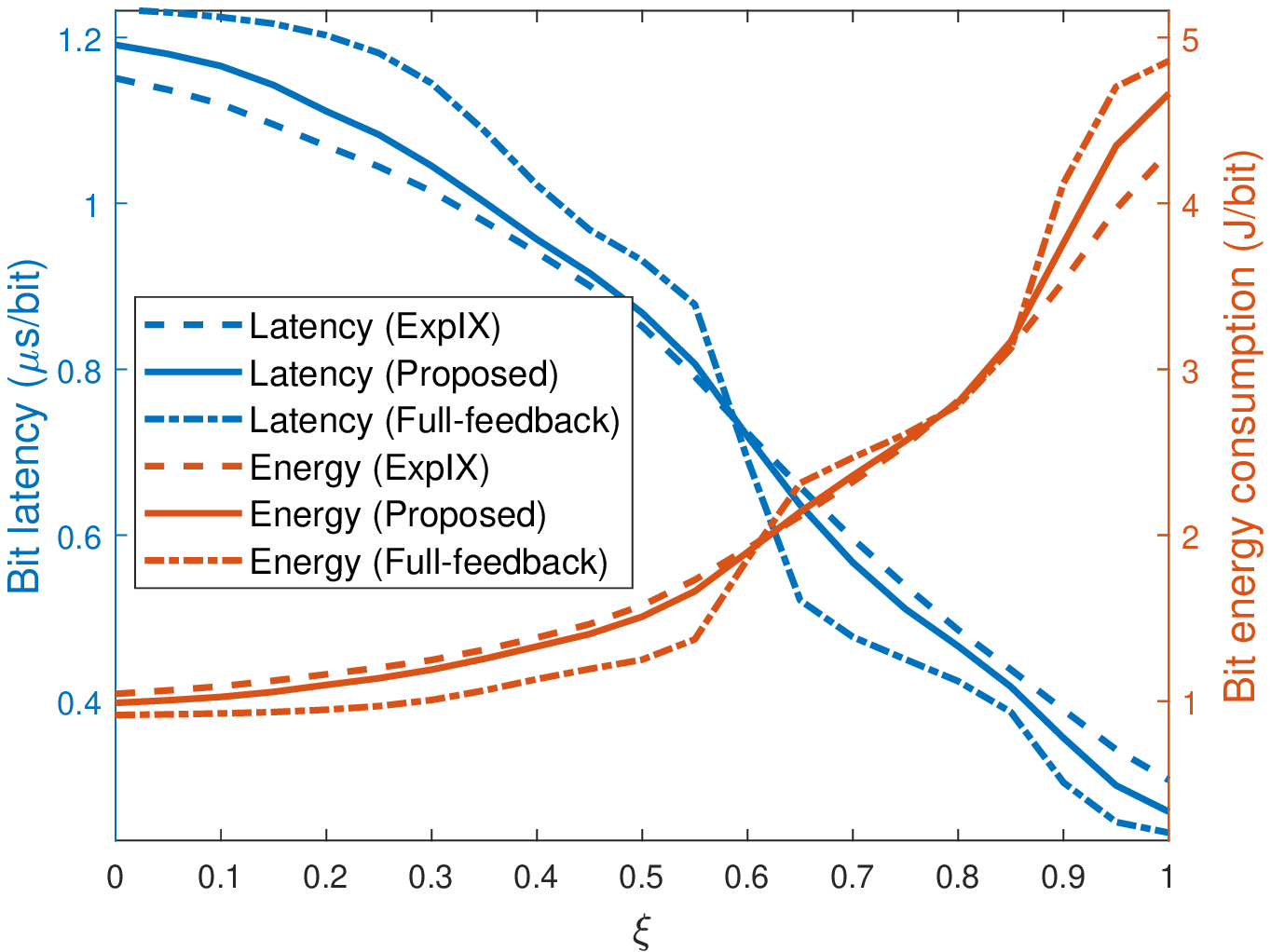}} & \hspace{-15pt} 
  		\subfigure[\label{numerical:ee_regret}]
		{\includegraphics[width=0.26\textwidth]{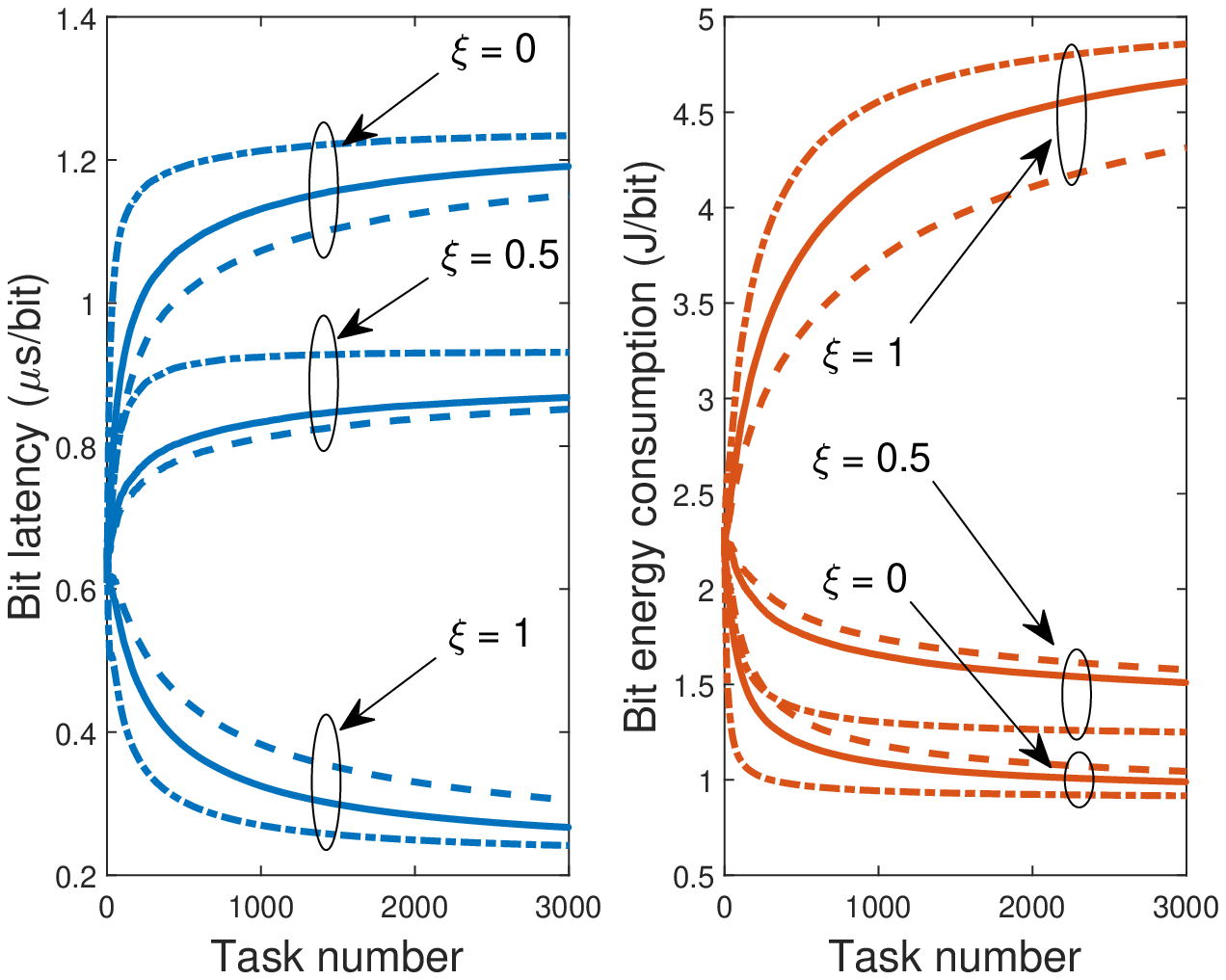}}
	\end{tabular}\vspace{-.1em}
	\caption{Impact of cost function weights on the average per-bit performance: (a) Per-bit cost w.r.t $\xi$ and (b) Per-bit cost w.r.t $T$}\vspace{-1em}
\end{figure}

\subsubsection*{\textbf{Impact of $\xi$}}
Fig. \ref{numerical:ee_latency} shows the impact of weighting parameter $\xi$ on the per-bit latency cost, $D_k^t/q^t$, and the per-bit energy consumption cost, $E_k^t/q^t$, of the proposed algorithm, vanilla Exp3IX algorithm and full-feedback case with $T = 3000$ tasks. It is noted that the proposed algorithm yields the per-bit cost values, latency and energy, each of which dwells between the individual per-bit cost values from Exp3IX and full feedback algorithm, in all $\xi$ regions. It is also observed that the individual per-bit cost of the proposed algorithm and that of the oracle fluidly move with respect to the weight parameter, $\xi$, i.e., smooth improvement or degradation. Increasing the weighting parameter makes the latency performance more dominated over another, and thus the per-bit latency performance is improved while the energy consumption performance gets less interesting. Such behavior can be captured in two different per-bit costs, latency and energy, in Fig. \ref{numerical:ee_regret}, when $\xi = 0, 0.5, 1$, respectively.
{To sum up, the effectiveness and robustness of the proposed algorithm are verified under the synthetic scenario, by showing its outperformance compared to other benchmarks in terms of the learning regret and the average per-bit cost, taking into account the dynamics of resource availability and demand in an adversarial environment. 
}

\subsection{Performance evaluation under realistic scenario} 
{ 
\subsubsection{Evaluation setting}
In this subsection, the applicability of the proposed task offloading algorithm is further explored by using the Luxembourg SUMO Traffic scenario (LuST) \cite{Codeca2015}. The Lust scenario simulates the real traffic in the
city of Luxembourg using SUMO, where arterial and residential roads link downtown and metropolitan areas with highways on the outskirts that surround the city. To better evaluate the resource supply volatility awareness of the proposed algorithm, we choose vehicle traffics on a highway road, e.g., consisting of multiple edges, IDs $31622\#5 \sim 31622\#10$, involved with multiple entrances and exits, as available VFNs. A task requester is assumed to have a full route on the highway, i.e., departing from edge ID $31622\#1$ every minute, and its candidate VFN set is volatile due to the facts: i) the VFNs may join or leave the highway, and ii) the vehicles move in the same direction at relatively fast but different speeds. The data of vehicle coordinate and velocity are used for simulation in MATLAB. The maximum CPU frequency of each VFN is randomly distributed in $[1, 5]$ GHz. The rest parameters follow the previous setting described for the synthetic scenario.

\begin{figure}[t]
	\centering
	\begin{tabular}{c c } \hspace{-20pt} 
		\subfigure[\label{numerical_higway:latency_8am}]
		{\includegraphics[width=0.27\textwidth]{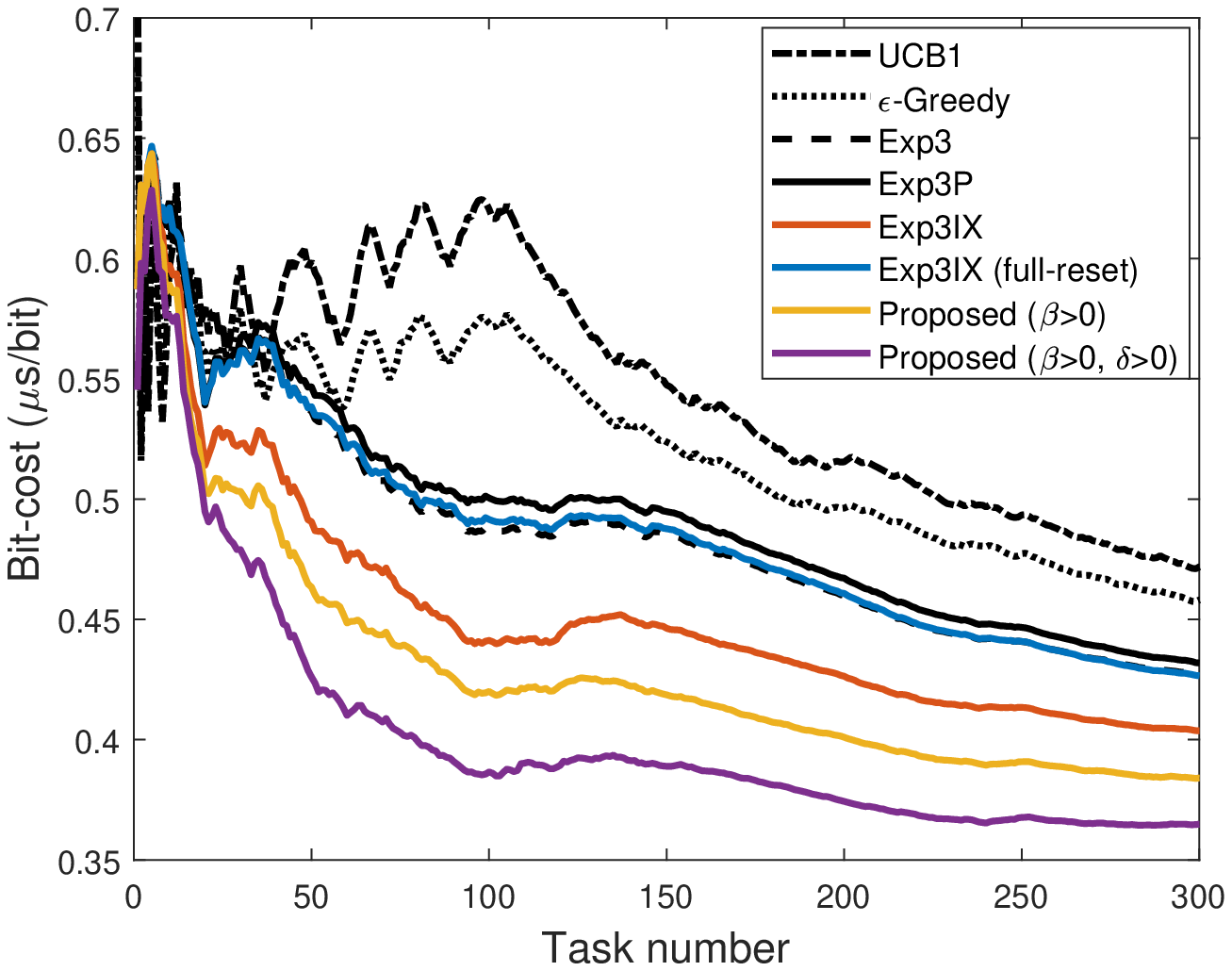}} &
		\hspace{-20pt}
		\subfigure[\label{numerical_higway:latency_1pm}]
		{\includegraphics[width=0.27\textwidth]{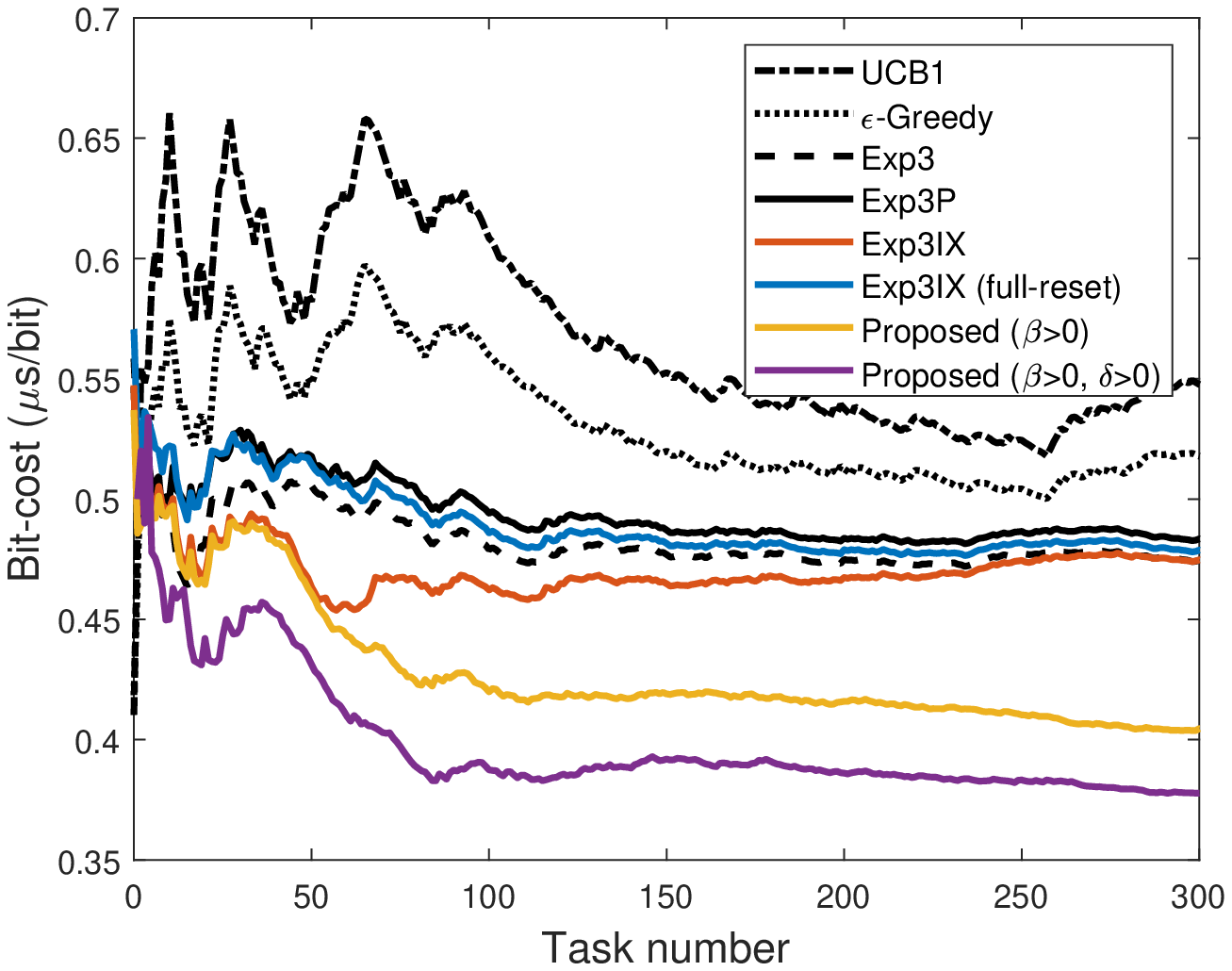}} \vspace{-0.5em}\\ \hspace{-20pt} 
		\subfigure[\label{numerical_higway:energy_8am}]
		{\includegraphics[width=0.27\textwidth]{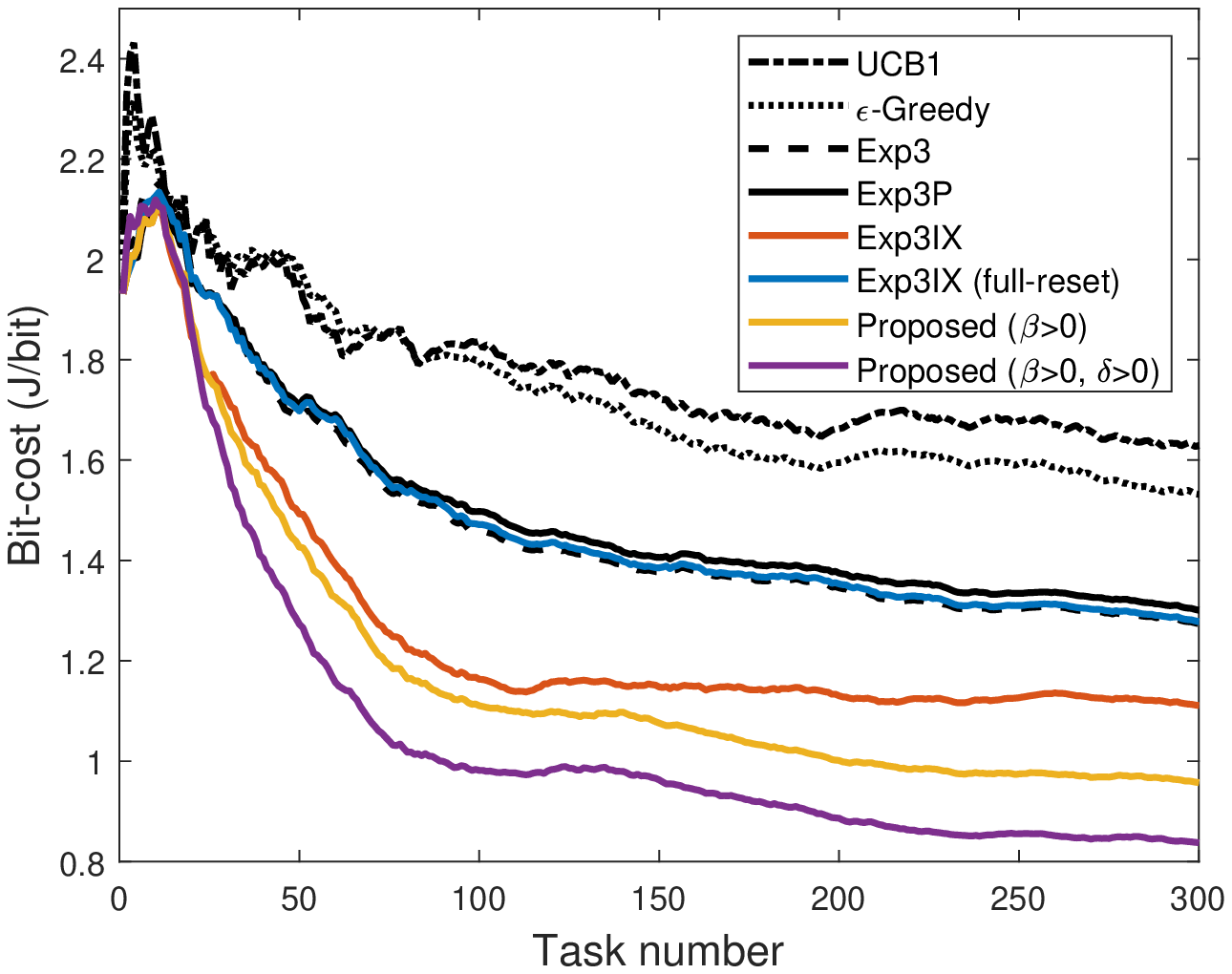}} &
		\hspace{-20pt}
		\subfigure[\label{numerical_higway:energy_1pm}]
		{\includegraphics[width=0.27\textwidth]{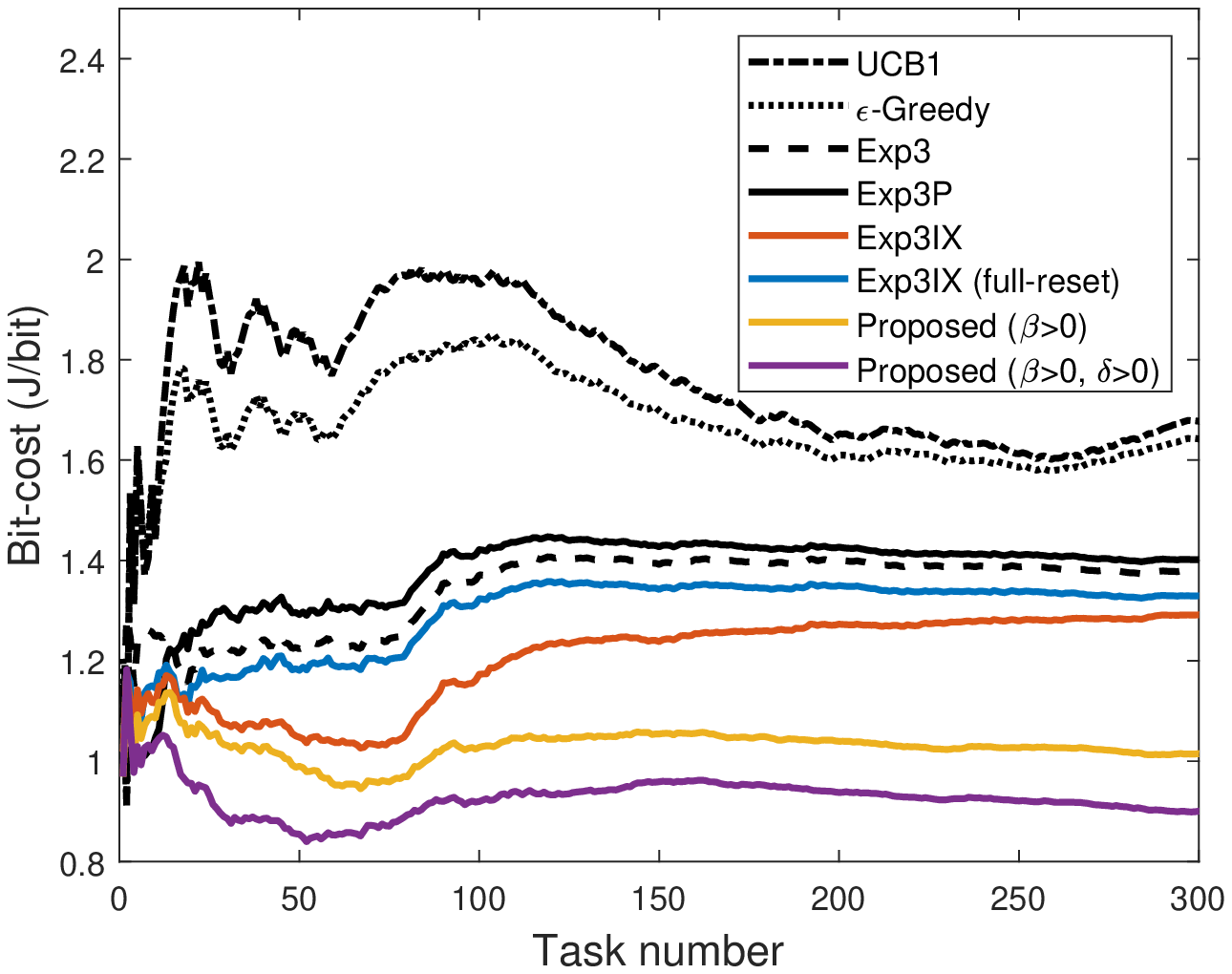}} \\
	\end{tabular}\vspace{-.1em}
	\caption{Average per-bit performance comparison with $\epsilon = 0.1$, under a highway scenario for different vehicle densities: (a) peak and (b) off-peak when $\xi = 1$, and (c) peak and (d) off-peak when $\xi = 0$. }\vspace{-1em} \label{numerical_higway}
\end{figure}


\subsubsection{Evaluation result}
Fig. \ref{numerical_higway} depicts the performance results of the proposed algorithm in terms of the average per-bit cost for different VFN densities. To consider different volumes of available VFNs, the scenario is running in two time windows, i) the morning rush hour peak period, $08:00\sim08:05$ and ii) the off-peak period around lunchtime, $13:00\sim13:05$ as in Fig. \ref{numerical_higway:latency_8am} and \ref{numerical_higway:latency_1pm}, respectively \cite{Zhu2018, Codeca2015}. The proposed task offloading algorithm always outperforms the other learning algorithms, since it can better adapt to the adversarial and dynamic environment better. To be specific, compared with the UCB and volatile Exp3IX algorithms, the proposed algorithm can reduce the average per-bit cost by about 23\% and 10\% (partial-reset) in peak time, see Fig. \ref{numerical_higway:latency_8am}, and  by about 30\% and 20\% (partial-reset) in off-peak time, see Fig. \ref{numerical_higway:latency_1pm}, when $T = 300$. The average per-bit cost decreases more at the expense of convergence rate for a high density of VFNs. The reason is that a large set of vehicles can extend the exploration space and increases the probabilities of finding good solutions, i.e., the average per-bit cost in peak time is more reduced than the one in off-peak. However, the larger exploration space tends to converge slower. Similar phenomena are also observed in per-bit energy cost in Fig. \ref{numerical_higway:energy_8am} and \ref{numerical_higway:energy_1pm}. 
} 

\section{Conclusions}
This work is to propose adaptive learning-based decentralized task offloading algorithm where each client can make the decision on fog node selection independently. The proposed online learning algorithm allows to provide the foundation for scalable and low-complexity offloading decision-making in an adversarial environment. In particular, two bottlenecks in the VFC-induced heterogeneous and dynamic environment, volatile candidate fog node set and task size, are addressed. We prove that the input-size dependent selection rule allows to choose a suitable fog node selection without exploring the sub-optimal actions, and also an appropriate score patching rule allows to quickly adapt to the evolving circumstance, thereby achieving better exploitation exploration balance. While this work focuses on self-interested regret-optimal, system-level perspective can be further considered, desirable to know whether the dynamical behaviors of distributed players promise a certain level of optimality in terms of social welfare under information limited case, i.e., unknown game.


\end{document}